\documentclass[pdflatex,sn-mathphys]{sn-jnl}

\jyear{2021}%

\usepackage{multirow}
\usepackage{hyperref}
\usepackage{booktabs}
\usepackage{amsmath}
\usepackage{subfigure}
\usepackage[switch]{lineno} 
\usepackage{natbib}

\begin{document}

\title[TED: RPT guided Tax Evasion Detection on Heterogeneous Graph]{TED: Related Party Transaction guided Tax Evasion Detection on Heterogeneous Graph}



\author[1]{\fnm{Yiming} \sur{Xu}}\email{xym0924@stu.xjtu.edu.cn}

\author*[1]{\fnm{Bin} \sur{Shi}}\email{shibin@xjtu.edu.cn}

\author[2]{\fnm{Bo} \sur{Dong}}\email{dong.bo@xjtu.edu.cn}

\author[1]{\fnm{Jiaxiang} \sur{Wang}}\email{wangjx@stu.xjtu.edu.cn}

\author[3]{\fnm{Hua} \sur{Wei}}\email{hua.wei@asu.edu}

\author[1]{\fnm{Qinghua} \sur{Zheng}}\email{qhzheng@mail.xjtu.edu.cn}

\affil[1]{\orgdiv{School of Computer Science and Technology}, \orgname{Xi’an Jiaotong University}, \orgaddress{\city{Shaanxi}, \country{China}}}

\affil[2]{\orgdiv{School of Distance Education}, \orgname{Xi’an Jiaotong University}, \orgaddress{\city{Shaanxi}, \country{China}}}


\affil[3]{\orgdiv{School of Computing and Augmented Intelligence}, \orgname{Arizona State University}, \orgaddress{\city{Phoenix}, \country{USA}}}

\abstract{
Tax evasion causes severe losses of government revenues and disturbs the economic order of fair competition.
To help alleviate this problem, the latest tax evasion detection solutions utilize expert knowledge to extract features and then train classifiers to determine whether a company is suspected of tax evasion.
However, existing solutions mainly focus on the statistical features of the company, but fail to exploit the rich interactive information in tax scenarios, which affect the detection performance. 
In this paper, we first model the tax scenario as a heterogeneous graph and study the tax evasion detection problem under the heterogeneous graph model. To improve the performance of tax evasion detection, a novel graph neural network model is proposed to extract the comprehensive information of heterogeneous graphs.
Specifically, we use heterogeneous and complex related party transaction groups to filter low-level noise information. Moreover, a hierarchical attention mechanism is designed to capture the deeper structure and semantic information hidden in the related party transaction group.
We apply our method to the real risk management system of the tax bureau, and evaluate it on two human-labeled real-world tax datasets. The results demonstrate that our method significantly outperforms the state-of-the-art in the tax evasion detection task. The code and data are available at: \url{https://github.com/yimingxu24/TED}.
}




\maketitle

\section{Introduction}
\label{sec:introduction}
With the rapid increase of market entities, the global economic competition is becoming increasingly fierce. To obtain more profits, the phenomenon of tax evasion has become more and more serious.
Tax evasion causes severe tax losses and undermines the fair competition of business environments.
Tax evasion is a global problem faced by almost all countries in the world. In the EU, the average share of the underground economy was 18.3\% in 2015 \cite{androniceanu2019interdependence}. 
Recent research \cite{lopez2017quantitative, cobham2018global, crivelli2015base} conservatively estimates that the global annual revenue loss is about  US\$500 billion. The loss of fiscal revenue is particularly serious in low-income and lower middle-income countries, which seriously obstructs the implementation of the public services and infrastructures.
The extent of tax evasion depends on the effectiveness of tax evasion detection by tax authorities \cite{allingham1972income}. Therefore, it is crucial to improve the effectiveness of tax evasion detection.

\begin{figure}
  \centering
  \includegraphics[width=8.8cm]{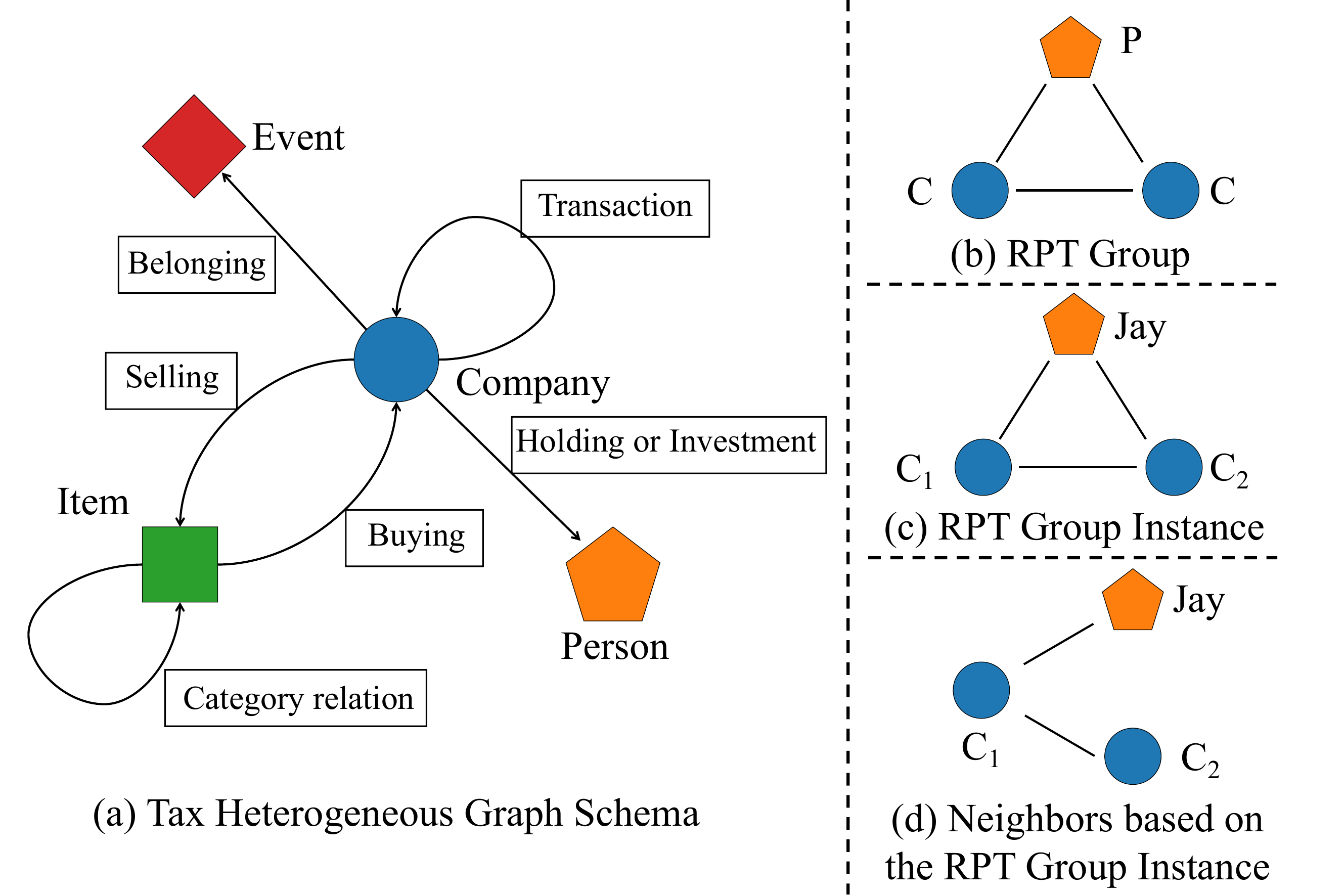}
  \caption{Heterogeneous graph of tax scenarios. (a) The schema of the heterogeneous graph, including four node types (i.e., company, person, item, event) and six edge types (i.e., transaction, holding or investment, selling, buying, belonging, and category relation). (b) The Person-Company-Company-Person ($PCCP$) RPT group. (c) An example instance of the $PCCP$ RPT group. (d) Instance neighbors of company $C_{1}$ (i.e., $Jay$, $C_{1}$ and $C_{2}$).}
  \label{fig:schema}
\end{figure}

In tax scenarios, there are a variety of relationships between taxpayers which could be very important for tax evasion detection. In recent financial area literatures~\cite{liu2017international, klassen2017transfer}, related party transaction (RPT) tax evasion is considered hard to be detected and has become a growing global trend. Through fake transactions between two related taxpayers, tax evasion can be easily achieved when these taxpayers hold a pre-existing relationship prior to the transaction~\cite{klassen2017transfer}. The relationships could be in a variety of forms. For example, in Fig.~\ref{fig:schema}(a), a company can be related to another company through the same person who controls both companies, or through buying/selling items in the same category. Therefore, it is necessary to investigate on the varied relationships between different parties.

In order to determine whether a company is suspected of tax evasion, nowadays tax authorities have been using feature-based methods, i.e., training classifiers on features manually selected by tax experts~\cite{didimo2020combining, rahimikia2017detecting, assylbekov2016detecting, stankevicius2015hybrid, de2018tax, savic2021tax}. However, the artificially designed features are difficult to portray the relationships between different parties, resulting in the loss of a large amount of interactive information. To better characterize the relationships, in this paper, we propose to use the heterogeneous graph to represent the parties and their relationships. As is shown Fig.~\ref{fig:schema}(a), the tax scenario could be represented as a heterogeneous graph, with four types of entities (company, person, item and event) and six relationships between them. To the best of our knowledge, this is the first study to tackle the tax evasion detection problem under the heterogeneous graph model. 

However, mining useful information from the heterogeneous graph in tax scenario is a non-trivial task with the following challenges: (1)~\textit{Noisy relationships}. There are many noisy relationships irrelevant to tax evasion detection in the heterogeneous graphs. How to reduce the negative impact of noisy nodes and edges in heterogeneous graphs is an intractable problem. (2)~\textit{Compound and covert relationships}. Different taxpayers could have different related parties, where certain taxpayer is more suspected of tax evasion (e.g., a person holding many companies might be more capable of tax evasion). Two taxpayers could also show different risks in tax evasion under different forms of relations, while their relationships could be covert (e.g., they are connected after multiple hops). How to differentiate and combine these influences is worth investigating.

To tackle the aforementioned challenges, in this paper, we first study the tax evasion detection problem under the heterogeneous graph model, and propose an RPT-guided Heterogeneous Graph Hierarchical Attention for \textbf{T}ax \textbf{E}vasion \textbf{D}etection, called TED. To tackle the challenge of noisy relationships, TED regards the RPT group as high-order proximity between nodes and subtly filters the RPT relationships out from other low-order noisy relationships. To tackle the challenge of compound relationships, a hierarchical attention framework is proposed to mine the more fine-grained and deeper structure and semantic information hidden in the RPT groups to enhance the feature representation of objects. Specifically, the hierarchical attention framework consists of two components: inner-RPT level and cross-RPT level. The inner-RPT level extracts features from each RPT around the node through the multiple head and attention mechanism.
And the cross-RPT level aggregates tax evasion features cross different RPT groups to learn the final node embeddings. Extensive experiments show the effectiveness of TED on the real-world tax datasets, where the F1 score and accuracy of TED are 8.50\% and 7.61\% higher than the best existing methods in the tax evasion detection task.

In summary, the main contributions are as follows:

$\bullet$ We propose to use heterogeneous graph for tax evasion detection. As far as we know, we are the first to study the tax evasion detection problem under the heterogeneous graph model with more than 10 kinds of nodes and relations. The comprehensive information of heterogeneous graphs can improve the performance of tax evasion detection.

$\bullet$
We propose to use the RPT group as high-order proximity to filter low-order noisy information, and propose a novel hierarchical attention model to learn the compound influences from heterogeneous, complex, and covert interactive relationships in tax scenarios.

$\bullet$
We apply our method to the real risk management system of the tax bureau in China, and evaluate it on two real-world tax datasets. The experimental results demonstrate that our method significantly outperforms the state-of-the-art methods in the tax evasion detection task. We also demonstrate the effectiveness of each component of our model through the ablation study, and visualize the attention weight of cross-RPT level to illustrate the importance of RPT for tax evasion detection.


\section{Related Work}
\label{sec:related_work}
\subsection{Tax Evasion Detection}
Tax evasion detection methods can be divided into three categories: rule-based methods, feature-based methods, and network-based methods~\cite{zheng2024survey}. 
Rule-based methods mainly use expert experience to build a rule-based system to screen abnormal financial indicators~\cite{liu2010application}.
However, the subjective judgment of experts makes the selection and update of rules expensive.

To solve this problem, researchers perform feature engineering from the inherent information of the company and use features to detect tax evasion. Pérez López et al.~\cite{perez2019tax} and Lin et al.~\cite{lin2015detecting} use neural network to detect tax evasion. González and Velásquez~\cite{gonzalez2013characterization} use clustering, decision trees, and neural networks to detect fraud patterns.
TEDM-PU~\cite{wu2019tedm} uses the random forest and LightGBM to train the model. 
However, these methods ignore the rich interactive information between companies, leading to a lack of information.

Therefore, scientific researchers try to use network-based methods to explore tax evasion in complex tax networks. Tian~\cite{tian2016mining} proposed a taxpayer benefit interaction network (TPIIN) to explore suspicious tax evasion groups. 
Didimo et al.~\cite{didimo2018visual,didimo2020combining} detect risky taxpayers through graph pattern matching and provide network visualization to tax analysts.
PUNE~\cite{mi2020tax} uses network embedding algorithms and PU learning to conduct tax evasion detection. Different from the existing methods, we use heterogeneous information networks to model multiple entity objects and relationship types in tax scenarios. 

\subsection{Graph Neural Networks} 
Recently, graph neural networks have received widespread attention~\cite{xu2025out,xu2025revisiting}. The graph neural network learns the low-dimensional embedding representation of each node through a message passing mechanism, which is used for a variety of downstream tasks~\cite{zhang2024graph，xu2025court}. Both spectral-based graph neural networks~\cite{defferrard2016convolutional, kipf2016semi} and spatial-based graph neural networks~\cite{hamilton2017inductive, xu2023cldg,xu2024learning} have been proposed and demonstrate powerful performance. 

All the above models are constructed for homogeneous graphs. However, the real-world complex network is often heterogeneous, including multiple object types and multiple link types. The main challenge is how to deal with heterogeneity to capture rich semantic information. 
Recently, graph neural network models based on heterogeneous graphs have been widely developed~\cite{dong2017metapath2vec, wang2019heterogeneous, fu2020magnn, hu2020heterogeneous,shi2023edge}. However, these methods are not designed to model tax scenarios, and they cannot effectively detect related party transaction tax evasion behaviors.


\section{Preliminaries}
\label{sec:preliminaries}
In this section, We first introduce the relevant definitions. Then, we elaborate on the motivation of this paper. 

\newcommand{\tabincell}[2]{\begin{tabular}{@{}#1@{}}#2\end{tabular}} 
\begin{table}[htbp]
  \centering
  \caption{Notations and Descriptions.}
  \label{tab:ND}
  \begin{tabular}{cc}
    \toprule
    Notations & Descriptions\\
    \midrule
    $G$ & A heterogeneous graph $G=\left ( V,E \right )$\\
    $V$ & The set of nodes in $G$\\
    $E$ & The set of edges in $G$\\
    $M$ & RPT group\\
    $a,\mathbf{a},\mathbf{A}$ & Scalar, vector, matrix\\
    $N_{i}^{M}$ & \tabincell{c}{Instance set of node $i$ based on RPT group $M$}\\
    $N_{ik}^{M}$ & \tabincell{c}{ The $k$th instance neighbor set of node $i$ based on RPT group $M$} \\
    
  \bottomrule
\end{tabular}
\end{table}

\subsection{Problem Formulation}
We list the definitions of some important terms in this paper and formally define the problem of tax evasion detection. Besides, we summarize the frequently used notations, 
which are illustrated in Table~\ref{tab:ND}.

\textbf{Definition 2.1. Heterogeneous Graph.} 
A heterogeneous graph is defined as a directed graph $G\left ( V,E\right )$ with a node type mapping function $\varphi:V\rightarrow A$ and a relation type mapping function $\psi :E\rightarrow R$, where each node $v\in V$ belongs to a specific object type in the object type set $A:\varphi \left ( v \right )\in A$, and each edge $e\in E$ belongs to a specific relationship type in the relation type set $R:\psi \left ( e \right )\in R$, respectively, with $\vert A \vert + \vert R \vert > 2$.

\textbf{Definition 2.2. RPT Group.} Formally, a RPT group is defined as a graph $M=\left ( V_{M},E_{M} \right )$, where $V_{M}$ is the set of nodes and $E_{M}$ is a set of edges in the RPT group $M$. For $\forall v\in V_{M}$ denotes a type from the node types $A$ and $\forall e\in E_{M}$ denotes a type from the edge types $R$, and any two company type nodes $u,v\in V_{M}$, $u,v$ must have a business relationship or common interest. As shown in Fig.~\ref{fig:schema}(b). 

\textbf{Definition 2.3. RPT Group Instance.} A graph $m=\left ( V_{m},E_{m} \right )$ is an instance of RPT group $M$ if there exists a function $f:V_{m}\rightarrow V_{M}$ between the set of nodes $V_{m}$ and $V_{M}$. For each node $v\in V_{m}$, $f\left ( v \right )=\varphi \left ( v \right )$ and for each node $u,v\in V_{m}$, edge $\left \langle u,v \right \rangle\in E_{m}$, $\left \langle f\left ( u \right ),f\left ( v \right ) \right \rangle\in E_{M}$ and $\left \langle f\left ( u \right ),f\left ( v \right ) \right \rangle=\psi \left ( \left \langle u,v \right \rangle \right )$. 

\textbf{Example.} Fig.~\ref{fig:schema}(c) shows two companies $C_{1}$ and $C_{2}$ that trade with each other are both invested by $Jay$. In this case, $C_{1}$ and $C_{2}$ could be easier to achieve tax evasion.

\textbf{Definition 2.4. Neighbors based on RPT Group.} Given a node $i$ and a RPT group $M$, $N_{ik}^{M}$ is defined as the set of nodes that connect with node $i$ through the $k$th RPT group instance of $M$, note that $N_{ik}^{M}$ includes node $i$. $N_{i}^{M}$ represents the instance set of node $i$ based on RPT group $M$.

\textbf{Example.} Take Fig.~\ref{fig:schema}(d) as an example. Given the RPT instance in Fig.~\ref{fig:schema}(c), the neighbors of $C_{1}$ are $Jay$, $C_{2}$ and $C_{1}$.

\textbf{Definition 2.5. Tax Evasion Detection Problem.} In the tax evasion detection problem, the transaction related behavior of taxpayers is modeled as a heterogeneous information network $G=\left ( V,E \right )$. Our purpose is to detect the company's tax evasion. We assign a label $y_{v}\in \left \{ 0,1 \right \}$ on a company $v \in V \vert V.type="company"$ to indicate whether it has tax evasion. Specifically, given the tax heterogeneous network $G=\left ( V,E \right )$ and training set $D=\left \{ \left ( v,y_{v} \right ) \right \}$, our tax evasion detection problem is to judge the probability of tax evasion for the companies in the test set.

\begin{figure}[htbp]
  \centering
  \includegraphics[width=8.0cm]{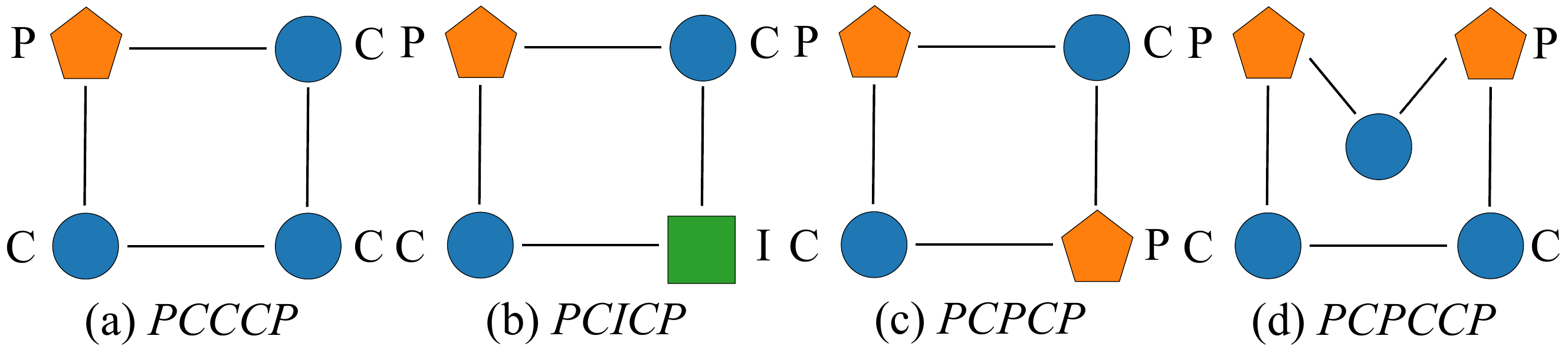}
  \caption{Four RPT groups. C, P, I nodes respectively represent the company, person and item node types, and the relationship between them is shown in Fig.~\ref{fig:schema}.}
  \label{fig:RPTschema}
\end{figure}

\begin{figure}[htbp]
  \centering
  \includegraphics[width=7.8cm]{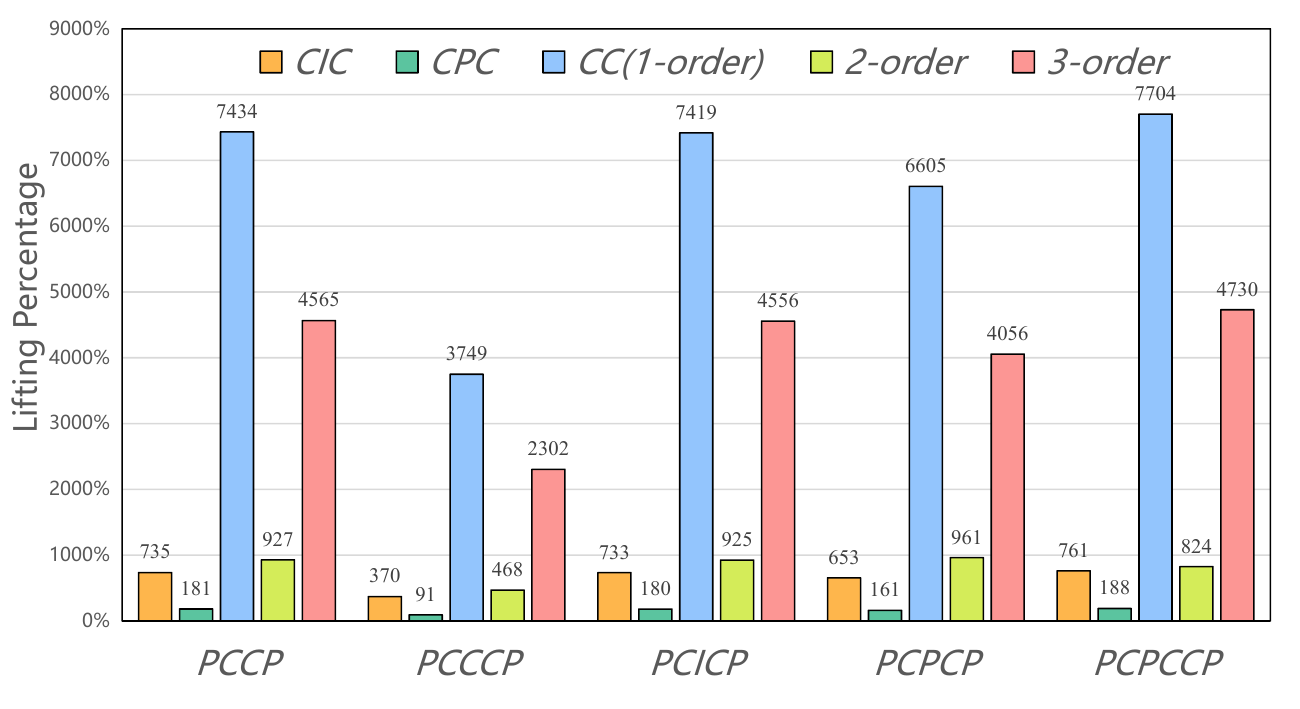}
  \caption{In the real-world dataset, the ratio of the probability that the RPT-group-based neighbors (i.e., $PCCP$, $PCCCP$, $PCICP$, $PCPCP$ and $PCPCCP$) of the tax evasion companies are also the tax evasion companies and the probability of the method based on the metapath neighbors (i.e., $CIC$, $CPC$,  $CC$) and the $k$-order neighbors ($k\leq 3$).}
  \label{fig:times}
\end{figure}

\subsection{Analysis of RPT}
We analyze tax evasion cases with experienced tax experts. We find that tax evasion is often more likely to occur in related party transaction groups, and the related tax evasion companies show compound relationships in the network, which is also a complex tax evasion behavior that is difficult to capture by the traditional linear neighbor selection method. As mentioned in the introduction, two companies are related parties if they have common beneficial owners, and companies connected through an RPT usually have a similar risk of tax evasion. For example, two companies with common investors, common business scope and trading behavior have more similar tax evasion risks than any other companies. To choose a neighbor selection method more suitable for tax heterogeneous information networks, with the help of experienced tax officers in the tax administration office, from the perspective of tax evasion detection, five RPT groups (as shown in Fig.~\ref{fig:schema}(b) and Fig.~\ref{fig:RPTschema}) and three metapaths ($CIC$, $CPC$ and $CC$, we refer you to~\cite{sun2011pathsim} for detailed notation and definition.) were designed, which can reflect the related interest relationship between tax evasion groups in different way. 

To intuitively illustrate the effectiveness of the RPT group as high-order neighbor, we calculate the probability that the RPT-group-based neighbors of the tax evasion companies are also tax evasion companies. Meanwhile, we also calculate the probability that the metapath neighbors and $k$-order neighbors of tax evasion companies are also tax evasion companies. Finally, we compared the probabilities\footnote{The results are based on the T20H dataset. The T15S dataset has similar results.}. The ratio is shown in Fig.~\ref{fig:times}:

$\bullet$
RPT-group-based neighbors of the tax evasion companies are more likely to be tax evasion companies.
The tax evasion probability of RPT group neighbors is about 200\% to 7000\% higher than that of metapath and $k$-order neighbor methods.
Compared with other neighbor selection methods, RPT group contains more tax evasion information and can filter low-order noise information.
Therefore, we need to use the RPT group to guide tax evasion detection.

$\bullet$
The neighbors of tax evasion companies based on different RPT groups also have different probabilities of tax evasion.
In essence, each RPT group contains different topology and compound relationships. 
Therefore, a method that can distinguish different RPT groups and mine deeper information from each RPT group is urgently needed.

\section{Proposed Method}
\label{sec:proposed_method}
\begin{figure*}
  \centering
  \includegraphics[width=12cm]{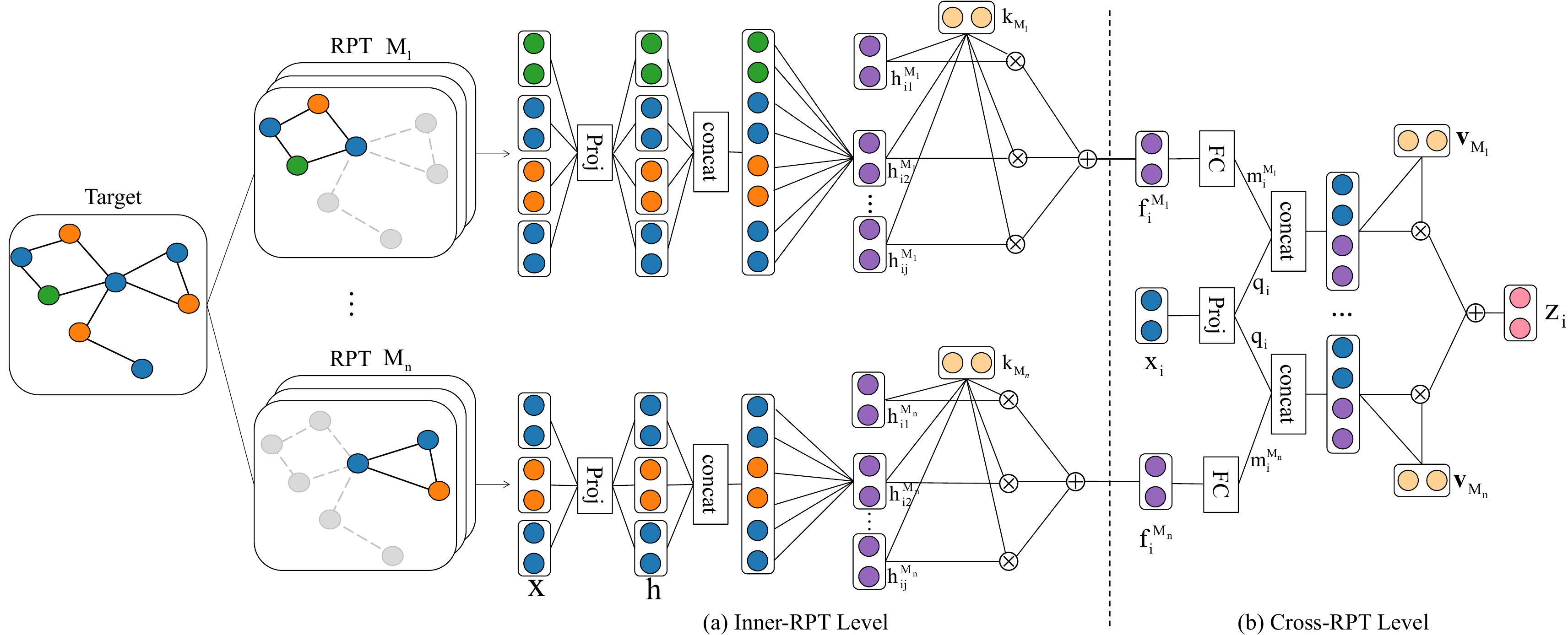}
  \caption{The overview of the TED. TED uses the RPT group as high-order proximity to filter low-order noisy relationships. Each RPT group contains different topology and compound relationships. Therefore, we first use the inner-RPT level to extract features from each RPT around the node, and then use the cross-RPT level to aggregate features cross RPT groups.}
  \label{fig:overview}
\end{figure*}

In this section, we propose a new graph neural network algorithm, namely TED, to make full use of the rich structure and tax evasion information brought by the RPT.
Fig.~\ref{fig:overview} shows an overview of TED, a hierarchical attention model that mainly includes the inner-RPT level, and the cross-RPT level.

\subsection{Inner-RPT Level}
There are many noise interactions in heterogeneous graphs that are not relevant to our problem. 
TED cleverly uses the RPT relationship, which is more important for tax evasion detection, while filtering out low-level noise information. 
For a given node, the inner-RPT level mainly extracts the tax risk information of each RPT group around it.
Each kind of RPT group contains multiple instances. Therefore, the inner-RPT level first needs to learn the feature of each RPT instance.

We use subgraph matching to select all RPT group instances. Each RPT group instance is heterogeneous and contains multiple types of nodes. Each type of node (e.g., company, person, item, and event) contains completely different attribute information, which has different feature spaces and may even have different feature dimensions.
Therefore, we cannot use them directly.
For each type of node, we design a corresponding feature transformation matrix to project different types of nodes into the same feature space. 
The projection process of node $i$ is as follows:
\begin{equation}
\mathbf{h}_{i}=\mathbf{P}_{\varphi \left ( i \right )}\cdot \mathbf{x}_{i},
\end{equation}
where $\mathbf{x}_{i}$ is the original feature vector and $\mathbf{h}_{i}$ is the feature vector after projection. $\mathbf{P}_{\varphi \left ( i \right )}$ is the projection transformation matrix of node type $\varphi \left ( i \right )$.

After the initial feature vector of the node is projected by the transformation matrix, we consider how to learn the feature of the RPT group instance. 
In the metapath based methods, researchers often discard all intermediate nodes in a metapath and convert the complex heterogeneous graph aggregation problem into a simple homogeneous graph problem, but this undoubtedly leads to considerable information loss. Therefore, we retain all neighbors of RPT group instances to reduce the loss of information. 
Specifically, we design a common instance feature transformation matrix for all instances in each RPT group, because they have the same topology.
The learning process of the instance $N_{ik}^{M}$ of node $i$ is as follows: 
\begin{equation}
\label{eq:rpt}
\mathbf{h}_{ik}^{M}=\sigma \left ( \mathbf{W}_{M}\cdot \left [ \mathbf{h}_{i}\overset{N_{ik}^{M}}{\underset{v\neq i}{\vert \vert}}\mathbf{h}_{v} \right ] \right ),
\end{equation}
where $\mathbf{h}_{i}$ and $\mathbf{h}_{v}$ are projection vectors, $N_{ik}^{M}$ is the $k$th instance neighbor set of node $i$ based on RPT group $M$, and $v\in N_{ik}^{M}$. $\vert \vert$ denotes the concatenate operation, $\mathbf{W}_{M}$ is the instance feature conversion matrix specific to the RPT. $\sigma \left ( \cdot  \right )$ is a nonlinear activation function and $\mathbf{h}_{ik}^{M}$ is the feature learned from node $i$ based on the $k$th instance of RPT group $M$. 

In addition, in order to better extract information, we use a multiple heads mechanism. The expression is as follows:
\begin{equation}
\mathbf{h}_{ik}^{M}=\overset{H}{\underset{h=1}{\vert \vert}}\sigma \left ( \mathbf{W}_{M}^{h}\cdot \left [ \mathbf{h}_{i}\overset{N_{ik}^{M}}{\underset{v\neq i}{\vert \vert}}\mathbf{h}_{v} \right ] \right ),
\end{equation}
where $H$ is the number of heads, and attention mechanisms perform Eq.~\ref{eq:rpt} $H$ times independent learning processes. The features are subsequently concatenated to obtain the resulting output features. $\mathbf{W}_{M}^{h}$ is the $h$th instance feature conversion matrix.

Given an RPT group, after extracting the RPT instance features, we should notice that each node contains multiple instances based on the RPT group, and each instance has different risk feature. Inspired by the attention mechanism, we introduced an attention mechanism to learn the importance of different instances in the RPT group. 
The importance of the $j$th instance $N_{ij}^{M}$ to node $i$ based on RPT group $M$ can be expressed as follows:
\begin{equation}
e_{ij}^{M}=\sigma \left ( \mathbf{k}_{M}^{T}\cdot \mathbf{h}_{ij}^{M} \right ),
\end{equation}
where $\mathbf{k}_{M}$ is the parameterized attention vector, which is specific to the RPT group type. $e_{ij}^{M}$ represents the importance of instance $N_{ij}^{M}$ to node $i$.
Then, normalize $e_{ij}^{M}$ using the softmax operation:
\begin{equation}
\alpha _{ij}^{M}=\frac{exp\left ( e_{ij}^{M} \right )}{\sum _{k\in N_{i}^{M}}exp\left ( e_{ik}^{M} \right )}, 
\end{equation}
where $\alpha _{ij}^{M}$ is the importance of the $j$th instance of the RPT group $M$ to the node $i$ after normalization, i.e., $\sum_{k}^{\vert N_{i}^{M} \vert}\alpha _{ik}^{M}=1$.

Then, the embedding of node $i$ based on the inner-RPT level can be aggregated by a set of instance embedding representations with the corresponding attention values as follows:
\begin{equation}
\mathbf{f}_{i}^{M}=\sigma \left ({\underset{j\in N_{i}^{M}}{\sum} \alpha _{ij}^{M}\cdot \mathbf{h}_{ij}^{M}} \right ),
\end{equation}
where $\mathbf{f}_{i}^{M}$ is the embedding representation learned by node $i$ based on RPT group $M$. 

\subsection{Cross-RPT Level}
After the inner-RPT level, each node obtains a set of RPT group features, i.e.,  $\left \{ \mathbf{f}_{i}^{1},\mathbf{f}_{i}^{2},\cdots ,\mathbf{f}_{i}^{\vert M \vert} \right \}$.
Each RPT group contains completely different topology and compound relationships in a heterogeneous graph. 
To better detect tax evasion, we propose a new cross-RPT level attention mechanism to mine more fine-grained and deeper information cross RPT groups. 

We first perform a nonlinear transformation on the inner-RPT level embedding, which is shown as follows:
\begin{equation}
\mathbf{m}_{i}^{M}=\sigma \left ( \mathbf{W}\cdot \mathbf{f}_{i}^{M} + \mathbf{b} \right ),
\end{equation}
where $\mathbf{W}$ is the weight matrix and $\mathbf{b}$ is the bias vector. 

Then, we perform a nonlinear transformation on the feature $x_{i}$ of the original node $i$, which will then be used as a part of the query vector in order to obtain a more accurate attention coefficient. This process is as follows:
\begin{equation}
\mathbf{q}_{i}=\sigma \left ( \mathbf{Q}\cdot \mathbf{x}_{i} \right ),
\end{equation}
where $\mathbf{Q}$ is the query matrix and $\mathbf{q}_{i}$ is the feature vector of node i after linear transformation. 

Each type of RPT group contains different structures and compound relationships, and their influence on nodes is quite different. 
We concatenate the nonlinear transformed feature vector $\mathbf{q}_{i}$ and the inner-RPT level embedding to obtain the query vector of node $i$. 
Then, we use the query vector and the RPT group specific attention vector $\mathbf{v}_{M}$ to measure the importance of each RPT group to the node, which is shown as follows:
\begin{equation}
e_{i}^{M}=\sigma \left ( \frac{\mathbf{v}_{M}^{T}\cdot \left [ \mathbf{q}_{i}\vert \vert \mathbf{m}_{i}^{M} \right ]}{\sqrt{d}} \right ),
\end{equation}
where $\mathbf{v}_{M}$ is the attention vector specific to the RPT group type $M$. $d$ is the output dimension of the node, and $e_{i}^{M}$ is a measure of the importance of RPT group $M$ to node $i$. 

Then, normalize $e_{i}^{M}$ using the softmax operation:
\begin{equation}
\beta _{i}^{M_{j}}=\frac{exp\left ( e_{i}^{M_{j}} \right )}{\sum_{k=1}^{\vert M \vert}exp\left ( e_{i}^{M_{k}} \right )},
\end{equation}
where $\beta _{i}^{M_{j}}$ is the normalized importance coefficient of the RPT group $M_{j}$ to node $i$, i.e. $\sum_{j=1}^{\vert M \vert}\beta _{i}^{M_{j}}=1$.

Finally, the learned attention coefficient $\beta _{i}$ is used to aggregate the specific embedding of the RPT group to obtain the final embedding $z_{i}$ as follows:
\begin{equation}
\mathbf{z}_{i}=\sum_{j=1}^{\vert M \vert}\beta _{i}^{M_{j}}\cdot \mathbf{m}_{i}^{M_{j}},
\end{equation}
where $\mathbf{z}_{i}$ is the embedding representation finally learned by node $i$, which can be used tax evasion detection task. 

\subsection{Objective and Model Training}
After the cross-RPT level, we obtain the final node representations, which are used for tax evasion detection tasks. 
We adopt a semi-supervised learning method, and define the tax evasion detection task as a classification task.
It is natural to minimize the cross-entropy loss between the prediction result and the actual label to optimize the embedding of the node:
\begin{equation}
L=-\sum_{y_{v}\in \mathbf{Y}}y_{v}log\left ( p_{v} \right ) +\left ( 1-y_{v} \right )log\left ( 1-p_{v} \right ),
\end{equation}
where $y_{v}\in \left \{ 0,1 \right \}$ is the label value of node $v$, and $p_{v}$ is the probability that node $v$ is predicted to be a tax evasion company. 

\begin{table}[htbp]
\centering
  \caption{Statistics of the Datasets.}
  \label{tab:SD}
  \begin{tabular}{cccccccc}
    \toprule
    Dataset & node type & node & edge type & edge & attribute \\
    \midrule
    T20H & 4 & 112015 & 6 & 198903 & 300 \\
    T15S & 2 & 132522 & 2 & 467273 & 300 \\
    \bottomrule
  \end{tabular}
\end{table}

\begin{figure}[htbp]
\centering
\subfigure[T20H]{
\includegraphics[width=5.2cm]{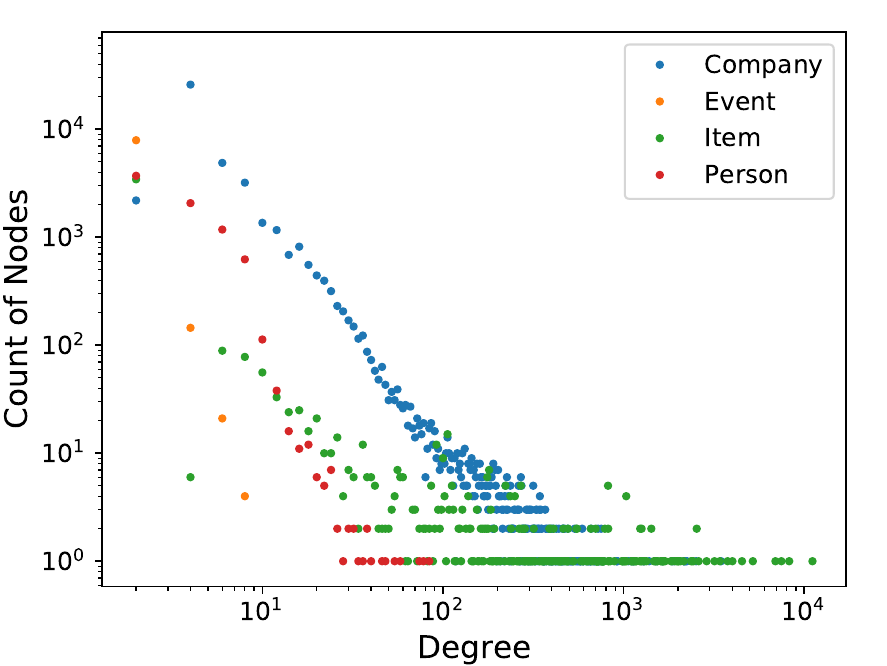}
}
\subfigure[T15S]{
\includegraphics[width=5.2cm]{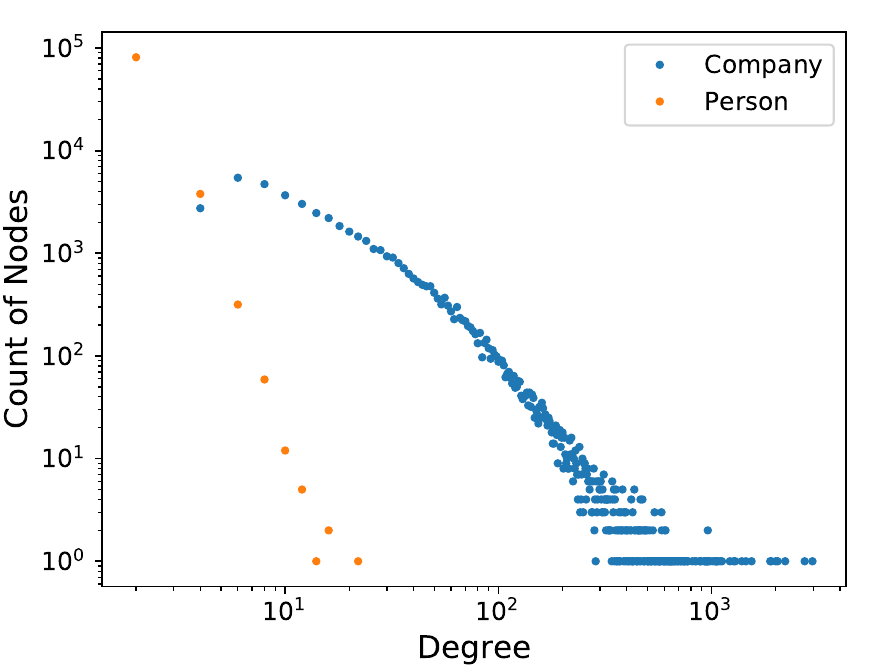}
}
\caption{Degree distribution in two real-world tax datasets.}
\label{fig:datasets}
\end{figure}

\section{Experiments}
\label{sec:experiments}
In this section, we first introduce two real-world tax datasets. 
Then, we conduct extensive experiments to prove the effectiveness of the proposed model. 
Finally, we perform robustness analysis, sensitivity study, ablation study, visualization analysis, attention trend analysis, efficiency and scalability and generality analysis.

The experiment aims to answer the following questions:

$\bullet$
$\mathbf{RQ_{1}}$: Is TED outperforming all state-of-the-art baselines for tax evasion detection?

$\bullet$
$\mathbf{RQ_{2}}$: How does TED perform in robustness?

$\bullet$
$\mathbf{RQ_{3}}$: How do various hyper-parameters impact TED performance?

$\bullet$
$\mathbf{RQ_{4}}$: How do different components affect TED performance?

$\bullet$
$\mathbf{RQ_{5}}$: How do we intuitively understand the representation capability of TED?

$\bullet$
$\mathbf{RQ_{6}}$:How can we intuitively understand the importance of RPT in tax evasion detection?

$\bullet$
$\mathbf{RQ_{7}}$: How does TED perform in efficiency and scalability?

$\bullet$
$\mathbf{RQ_{8}}$: Can TED be applied in different countries and regions?

\subsection{Experiment Preparation}
\subsubsection{Real-world Tax Dataset}
\ 
\newline
We construct two tax heterogeneous graph datasets T20H and T15S from the data provided by two tax bureaus.
The statistics of the two datasets are shown in Table~\ref{tab:SD}, and the degree distribution is shown in Fig.~\ref{fig:datasets}. 
We define the positive samples as tax evasion companies and the negative samples as normal companies. 
Since labeled samples are very difficult to get in real tax scenarios, many companies in the datasets are with no label. For this reason, we use a semi-supervised method to train the model.
The labeled samples were divided into training and test sets, with 1770 (3.99\%), 398 (0.90\%) in T20H and 2072 (4.41\%), 514 (1.09\%) in T15S, respectively. To simulate the real tax evasion scenarios where the proportion of tax evasion companies are unknown, 
we further set 5 control groups, changing the positive samples ratio (PSR) of traning sets as 50\%, 40\%, 30\%, 20\%, and 10\%.

\subsubsection{Baselines}
\ 
\newline
To verify the effectiveness of the model, we compare our model with some state-of-art baselines in tax evasion detection task. For a fair comparison, except the PU learning based models (which cannot output node embedding), other models output the embedding representation of nodes after training, and then uniformly input them into the classifier to evaluate the performance of the model. 

\textbf{Proximity-Preserving based Methods:}

\noindent$\bullet$
\textbf{HIN2Vec}~\cite{fu2017hin2vec}: The model first generates data based on random walks and negative sampling. It maximizes the possibility of predicting the relationship between nodes.

\noindent$\bullet$
\textbf{PTE}~\cite{tang2015pte}: The model was modified from the original implementation to allow the model to directly use heterogeneous networks with any type of nodes and links. 

\noindent$\bullet$
\textbf{Metapath2vec}~\cite{dong2017metapath2vec}: The model was modified first to perform a random walk to learn the weights of different metapaths based on the number of sampled instances and then use a unified loss function to train the model.

\textbf{Relation-Learning based Methods:} 

\noindent$\bullet$
\textbf{TransE}~\cite{bordes2013translating}: The method models relationships by interpreting them as translations operating on the low-dimensional embeddings of the entities.

\noindent$\bullet$
\textbf{ConvE}~\cite{dettmers2018convolutional}: ConvE is a representation learning model using a 2D convolutional neural network. The model was modified to only output node embeddings. 

\noindent$\bullet$
\textbf{DistMult}~\cite{yang2014embedding}: DistMult is a simple formulation of the bilinear model. The model was modified and only outputs node embedding.

\noindent$\bullet$
\textbf{ComplEx}~\cite{trouillon2016complex}: In ComplEx, by introducing the complex number space. The model was modified and only outputs node embedding.

\textbf{PU Learning based Methods:} 

\noindent$\bullet$
\textbf{TEDM-PU}~\cite{wu2019tedm}: TEDM-PU uses the LightGBM~\cite{ke2017lightgbm} method for tax evasion detection.

\noindent$\bullet$
\textbf{PUNE}~\cite{mi2020tax}: PUNE uses network embedding algorithms and PU learning for tax evasion detection.

\textbf{Message-Passing based Methods:} 

\noindent$\bullet$
\textbf{GCN}~\cite{kipf2016semi}: GCN stimulates the choice of convolution structure by the local first-order approximation of spectral convolution.

\noindent$\bullet$
\textbf{HAN}~\cite{wang2019heterogeneous}: HAN uses both node-level attention and semantic-level attention to aggregate information from a homogeneous graph based on metapaths. 

\noindent$\bullet$
\textbf{HGT}~\cite{hu2020heterogeneous}: The model aggregates information from the source node through heterogeneous mutual attention, heterogeneous message passing, and aggregation for specific tasks and obtains the context of the target node. 

\noindent$\bullet$
\textbf{R-GCN}~\cite{schlichtkrull2018modeling}: R-GCN applies parameter sharing and sparse constraint models to multiple graphs with many relationships.

\subsubsection{Implementation Details}
\ 
\newline
For our TED model, we use the Adam optimizer~\cite{kingma2014adam}, the learning rate is 0.005, the weight decay is 0.0005, the batch size is 256, the number of heads to extract the features of the RPT group instance is 8 and 6, respectively at T20H and T15S, and the $\alpha $ in LeakyReLU is 0.2. 
For TEDM-PU, the model uses only the initialized embedding of company nodes for tax evasion detection and does not use network information. For homogeneous graph methods PUNE and GCN, we only use nodes of the company type and the edges between them. We choose the same metapaths (as mentioned in 3.2) designed from the perspective of tax evasion detection for all the methods that need to clearly define the metapath. 
For the baselines, the implementation details can refer to the original paper or HNE~\cite{yang2020heterogeneous} and we optimize their parameters according to their paper. We set the embedding dimension of all the above algorithms to 50.

\begin{sidewaystable}[htbp]
\centering
\setlength\tabcolsep{1pt}
\caption{Experimental results (\%) of tax evasion detection task. TED outperforms baselines.}
\resizebox{\textwidth}{!}{
\begin{tabular}{|c|c|c|ccc|cccc|cc|ccccc|}
\hline
\multicolumn{1}{|c|}{\multirow{2}{*}{Datasets}} & \multicolumn{1}{c|}{\multirow{2}{*}{Metrics}} & \multicolumn{1}{c|}{\multirow{2}{*}{PSR}} & \multicolumn{3}{c|}{Proximity-Preserving based}                                                                      & \multicolumn{4}{c|}{Relation-Learning based}                                                                                                  & \multicolumn{2}{c|}{PU Learning based}                                   & \multicolumn{5}{c|}{Message-Passing based}                                                                                                                   \\ \cline{4-17} 
\multicolumn{1}{|c|}{}                          & \multicolumn{1}{c|}{}                         & \multicolumn{1}{c|}{}                    & HIN2Vec & PTE  & metapath2vec & TransE & ConvE & DistMult & ComplEx & TEDM-PU & PUNE  & GCN   & HAN   & HGT & \scriptsize{R-GCN} & \scriptsize{TED} \\ \hline
\multirow{10}{*}{T20H} & \multirow{5}{*}{F1}  & 50\% & 84.09$_{\pm 1.18}$   & 82.50$_{\pm 1.34}$ & 73.73$_{\pm 3.10}$        & 79.51$_{\pm 2.18}$  & 87.24$_{\pm 1.40}$ & 87.76$_{\pm 1.18}$   & 80.63$_{\pm 3.55}$   & 81.18$_{\pm 2.83}$   & 89.84$_{\pm 1.92}$ & 75.69$_{\pm 1.65}$ & 89.15$_{\pm 0.90}$ & 88.84$_{\pm 1.81}$ & 80.26$_{\pm 1.82}$ & \textbf{94.96}$_{\pm 0.93}$ \\
                          &                      & 40\% & 81.79$_{\pm 1.73}$   & 86.10$_{\pm 0.87}$ & 77.55$_{\pm 2.19}$        & 84.85$_{\pm 1.49}$  & 85.56$_{\pm 1.65}$ & 88.19$_{\pm 1.51}$    & 88.38$_{\pm 1.53}$   & 79.91$_{\pm 2.40}$   & 89.11$_{\pm 1.69}$ & 78.37$_{\pm 1.51}$ & 91.84$_{\pm 0.94}$ & 89.95$_{\pm 1.26}$ & 81.69$_{\pm 1.90}$ & \textbf{93.83}$_{\pm 0.72}$ \\
                          &                      & 30\% & 79.33$_{\pm 1.67}$    & 86.18$_{\pm 1.28}$ & 59.93$_{\pm 2.07}$        & 84.40$_{\pm 1.59}$  & 83.29$_{\pm 1.26}$ & 87.70$_{\pm 2.03}$    & 87.37$_{\pm 1.63}$   & 76.30$_{\pm 3.19}$   & 87.96$_{\pm 1.91}$ & 76.30$_{\pm 0.86}$ & 86.98$_{\pm 0.88}$ & 87.36$_{\pm 0.97}$ & 74.58$_{\pm 1.82}$ & \textbf{92.78}$_{\pm 1.01}$ \\
                          &                      & 20\% & 69.97$_{\pm 1.94}$    & 86.10$_{\pm 1.14}$ & 53.33$_{\pm 2.74}$        & 77.49$_{\pm 2.14}$  & 83.80$_{\pm 1.89}$ & 83.98$_{\pm 1.30}$    & 86.03$_{\pm 2.43}$   & 64.59$_{\pm 2.28}$   & 85.78$_{\pm 0.96}$ & 62.99$_{\pm 2.43}$ & 75.16$_{\pm 0.98}$ & 78.24$_{\pm 1.14}$ & 54.29$_{\pm 1.60}$ & \textbf{91.43}$_{\pm 1.19}$ \\
                          &                      & 10\% & 59.65$_{\pm 2.79}$    & 88.27$_{\pm 1.58}$ & 47.83$_{\pm 2.79}$         & 83.71$_{\pm 2.96}$  & 79.64$_{\pm 2.00}$ & 78.42$_{\pm 1.83}$    & 74.61$_{\pm 2.67}$   & 77.69$_{\pm 2.92}$   & 84.53$_{\pm 1.41}$ & 58.42$_{\pm 1.89}$  & 63.05$_{\pm 1.20}$ & 69.23$_{\pm 1.69}$  & 38.71$_{\pm 2.86}$ & \textbf{91.73}$_{\pm 1.31}$ \\ \cline{2-17} 
                          & \multirow{5}{*}{Acc} & 50\% & 84.18$_{\pm 1.52}$   & 82.41$_{\pm 1.33}$ & 67.59$_{\pm 2.79}$        & 74.87$_{\pm 2.65}$  & 86.18$_{\pm 1.83}$ & 87.94$_{\pm 1.40}$    & 78.39$_{\pm 3.27}$   & 77.64$_{\pm 2.66}$   & 89.20$_{\pm 2.05}$ & 73.37$_{\pm 1.72}$ & 88.44$_{\pm 0.64}$ & 87.94$_{\pm 1.48}$ & 76.88$_{\pm 2.15}$ & \textbf{94.72}$_{\pm 0.83}$ \\
                          &                      & 40\% & 82.66$_{\pm 1.97}$   & 86.93$_{\pm 1.07}$ & 77.89$_{\pm 2.15}$        & 83.67$_{\pm 1.87}$  & 86.18$_{\pm 1.87}$ & 88.69$_{\pm 1.39}$    & 88.44$_{\pm 1.56}$   & 76.38$_{\pm 2.50}$   & 88.94$_{\pm 1.84}$ & 78.64$_{\pm 2.01}$ & 91.96$_{\pm 0.79}$ & 90.45$_{\pm 1.60}$ & 80.40$_{\pm 1.45}$ & \textbf{93.72}$_{\pm 0.90}$ \\
                          &                      & 30\% & 81.41$_{\pm 1.88}$   & 87.19$_{\pm 1.56}$ & 69.10$_{\pm 2.24}$        & 84.67$_{\pm 1.51}$  & 84.67$_{\pm 1.11}$ & 88.44$_{\pm 2.52}$    & 87.94$_{\pm 1.72}$   & 74.87$_{\pm 3.25}$   & 87.69$_{\pm 2.31}$ & 79.40$_{\pm 1.40}$ & 88.19$_{\pm 0.76}$ & 88.44$_{\pm 1.64}$ & 77.39$_{\pm 1.79}$ & \textbf{92.96}$_{\pm 0.95}$ \\
                          &                      & 20\% & 75.63$_{\pm 2.18}$   & 87.19$_{\pm 1.84}$ & 66.58$_{\pm 2.53}$        & 80.15$_{\pm 2.07}$  & 85.43$_{\pm 1.87}$ & 85.43$_{\pm 1.35}$    & 87.19$_{\pm 2.48}$   & 68.59$_{\pm 2.42}$   & 84.67$_{\pm 1.32}$ & 71.36$_{\pm 2.69}$ & 79.90$_{\pm 0.86}$ & 81.41$_{\pm 1.25}$ & 67.84$_{\pm 1.16}$ & \textbf{91.71}$_{\pm 0.89}$ \\
                          &                      & 10\% & 71.11$_{\pm 2.65}$   & 89.45$_{\pm 1.86}$ & 63.82$_{\pm 2.80}$         & 85.43$_{\pm 2.92}$  & 82.91$_{\pm 1.93}$ & 82.16$_{\pm 2.21}$    & 79.65$_{\pm 3.26}$   & 76.63$_{\pm 3.34}$   & 83.17$_{\pm 1.97}$ & 69.60$_{\pm 2.18}$ & 72.61$_{\pm 0.96}$ & 75.88$_{\pm 1.45}$ & 61.81$_{\pm 0.79}$ & \textbf{92.21}$_{\pm 1.07}$ \\ \hline
\multirow{10}{*}{T15S} & \multirow{5}{*}{F1}  & 50\% & 55.40$_{\pm 2.13}$   & 52.31$_{\pm 0.73}$ & 54.20$_{\pm 2.14}$         & 52.75$_{\pm 3.49}$  & 55.13$_{\pm 2.16}$ & 46.75$_{\pm 1.47}$    & 48.14$_{\pm 2.24}$   & 79.45$_{\pm 1.92}$   & 66.84$_{\pm 0.46}$ & 70.04$_{\pm 1.22}$ & 79.69$_{\pm 0.83}$ & 81.54$_{\pm 0.66}$ & 79.83$_{\pm 2.38}$ & \textbf{93.36}$_{\pm 0.65}$ \\
                          &                      & 40\% & 28.32$_{\pm 2.01}$    & 38.34$_{\pm 1.42}$ & 47.66$_{\pm 2.47}$        & 25.84$_{\pm 3.14}$  & 35.98$_{\pm 3.45}$ & 32.58$_{\pm 1.81}$    & 34.39$_{\pm 1.78}$   & 77.87$_{\pm 2.69}$   & 71.99$_{\pm 1.46}$ & 54.35$_{\pm 2.62}$ & 77.61$_{\pm 0.89}$ & 79.17$_{\pm 0.50}$ & 81.93$_{\pm 1.45}$ & \textbf{93.39}$_{\pm 0.73}$ \\
                          &                      & 30\% & 19.21$_{\pm 2.27}$    & 34.50$_{\pm 1.78}$  & 34.48$_{\pm 1.83}$         & 22.98$_{\pm 1.96}$   & 30.64$_{\pm 2.63}$  & 16.72$_{\pm 1.06}$     & 30.00$_{\pm 1.72}$    & 70.78$_{\pm 1.96}$   & 70.98$_{\pm 1.31}$ & 39.64$_{\pm 2.49}$ & 74.25$_{\pm 0.85}$ & 74.17$_{\pm 1.37}$ & 76.21$_{\pm 1.76}$ & \textbf{94.96}$_{\pm 0.93}$ \\
                          &                      & 20\% & 11.55$_{\pm 1.93}$    & 20.58$_{\pm 1.62}$  & 20.78$_{\pm 2.11}$         & 18.53$_{\pm 2.65}$   & 24.53$_{\pm 1.02}$  & 3.04$_{\pm 0.97}$     & 24.52$_{\pm 2.18}$    & 54.59$_{\pm 2.41}$   & 67.02$_{\pm 0.94}$ & 33.96$_{\pm 2.95}$  & 64.18$_{\pm 0.91}$ & 70.70$_{\pm 0.95}$ & 74.16$_{\pm 1.04}$ & \textbf{86.59}$_{\pm 0.98}$ \\
                          &                      & 10\% & 10.79$_{\pm 2.26}$     & 10.04$_{\pm 1.19}$  & 3.73$_{\pm 1.09}$         & 5.26$_{\pm 1.12}$   & 16.33$_{\pm 2.46}$  & 2.29$_{\pm 0.53}$     & 11.43$_{\pm 1.60}$    & 28.19$_{\pm 2.58}$   & 64.68$_{\pm 1.65}$ & 16.25$_{\pm 2.61}$  & 41.18$_{\pm 0.97}$ & 66.50$_{\pm 1.79}$ & 60.98$_{\pm 1.99}$ & \textbf{76.35}$_{\pm 1.12}$ \\ \cline{2-17} 
                          & \multirow{5}{*}{Acc} & 50\% & 57.39$_{\pm 2.69}$    & 51.75$_{\pm 0.80}$ & 53.31$_{\pm 2.55}$        & 51.56$_{\pm 3.97}$  & 54.00$_{\pm 2.89}$ & 52.14$_{\pm 2.09}$    & 48.44$_{\pm 2.16}$   & 79.57$_{\pm 1.32}$   & 50.19$_{\pm 0.95}$ & 71.21$_{\pm 1.53}$ & 79.57$_{\pm 0.72}$ & 81.32$_{\pm 1.13}$ & 81.52$_{\pm 3.19}$ & \textbf{93.19}$_{\pm 0.61}$ \\
                          &                      & 40\% & 52.72$_{\pm 2.24}$   & 48.05$_{\pm 1.17}$ & 50.00$_{\pm 2.27}$        & 55.03$_{\pm 3.34}$  & 52.92$_{\pm 3.97}$ & 53.31$_{\pm 1.96}$    & 51.75$_{\pm 1.99}$   & 78.99$_{\pm 2.32}$   & 62.45$_{\pm 1.99}$ & 66.34$_{\pm 2.67}$ & 79.57$_{\pm 0.76}$ & 80.54$_{\pm 0.96}$ & 80.35$_{\pm 1.00}$ & \textbf{93.39}$_{\pm 0.69}$ \\
                          &                      & 30\% & 52.53$_{\pm 2.89}$   & 54.72$_{\pm 1.93}$ & 51.95$_{\pm 2.12}$        & 51.75$_{\pm 1.89}$  & 53.31$_{\pm 2.76}$ & 51.56$_{\pm 1.25}$    & 53.70$_{\pm 1.69}$   & 75.10$_{\pm 2.48}$   & 63.42$_{\pm 1.64}$ & 60.89$_{\pm 2.35}$ & 76.65$_{\pm 0.75}$ & 77.24$_{\pm 1.26}$ & 78.02$_{\pm 1.25}$ & \textbf{94.94}$_{\pm 0.90}$ \\
                          &                      & 20\% & 52.33$_{\pm 2.11}$   & 51.95$_{\pm 1.91}$ & 52.53$_{\pm 2.42}$        & 50.39$_{\pm 2.49}$  & 53.31$_{\pm 2.12}$ & 50.39$_{\pm 1.16}$    & 54.47$_{\pm 2.10}$   & 66.34$_{\pm 3.06}$   & 51.75$_{\pm 1.17}$ & 59.14$_{\pm 2.98}$ & 71.98$_{\pm 0.81}$ & 76.46$_{\pm 1.17}$ & 77.63$_{\pm 0.58}$ & \textbf{87.16}$_{\pm 0.94}$ \\
                          &                      & 10\% & 51.75$_{\pm 2.36}$   & 51.17$_{\pm 1.51}$ & 49.81$_{\pm 1.37}$        & 50.97$_{\pm 2.06}$  & 52.14$_{\pm 2.38}$ & 50.19$_{\pm 1.28}$    & 51.75$_{\pm 1.95}$   & 57.39$_{\pm 3.19}$   & 56.23$_{\pm 1.91}$ & 53.89$_{\pm 3.01}$ & 61.09$_{\pm 0.89}$ & 74.32$_{\pm 1.67}$ & 70.62$_{\pm 1.01}$ & \textbf{80.35}$_{\pm 0.97}$ \\ \hline
\end{tabular}}
\footnotesize{*Note: The Metrics uses F1 score and accuracy. PSR is the positive sample ratio in the training set. }\\
\label{tab:ncTax20H}
\end{sidewaystable}

\subsection{Tax Evasion Detection Task ($\mathbf{RQ_{1}}$)}
To answer $\mathbf{RQ_{1}}$, we compare with 13 state-of-the-art algorithms on the two tax datasets. 
To simulate the real tax scenarios, we construct five different training datasets by dividing different positive sample ratios (PSR) to conduct sufficient tax evasion detection experiments.
For fairness, we use the embeddings learned on the training set nodes to train a separate SVM, and use the test set to evaluate the tax evasion detection performance of the model.
We repeat this process 5 times and use the average F1 score and accuracy as the evaluation metrics. 

The experimental results are shown in Table~\ref{tab:ncTax20H}. We can find that TED outperforms baselines. The T15S dataset has fewer node types than the T20H dataset, and more edges bring more noise information irrelevant to tax evasion. As a result of the T15S dataset, especially when the positive samples ratio is very low, the proximity-preserving based and relation-learning based methods predict most companies as negative samples, resulting in poor performance.
Generally, the graph neural network based methods combine structure and feature information and usually perform better and more stable in the two datasets. Compared with GCN, the heterogeneous graph neural network based models perform better, indicating that the rich information from heterogeneous information networks can improve the performance of tax evasion detection. The complex structure of the heterogeneous network makes the data processing and semantic mining process more difficult. This is why GCN outperforms some heterogeneous graph models in some cases. Existing semantic exploration methods for graph neural networks are mainly based on metapath, random walk and k-order neighbors. However, the neighbors obtained by these methods either have no actual meaning or contain less tax risk information (as introduced in Section 3.2). The reason for the better performance of TED is to regard RPT groups as higher-order neighbors between nodes, distinguish low-order noisy relationships, and subsequently mine deep information from RPT groups using hierarchical attention model. By visualizing the attention value at the cross-RPT level in Section 5.7, the importance of the RPT group for tax evasion detection is further verified.

Compared with the sub-optimal, in the T20H and T15S datasets, the F1 score increased by 4.14\% and 12.86\% on average, and the accuracy increased by 3.82\% and 11.40\% on average, respectively. In summary, heterogeneous and complex RPT groups in tax heterogeneous graphs can better shield low-order noise information, and the RPT group guided hierarchical attention model proposed in this paper outperforms all the state-of-the-art baselines on tax evasion detection.

\subsection{Robustness analysis ($\mathbf{RQ_{2}}$)}
As shown in Table~\ref{tab:ncTax20H}, TED achieves the best performance under five different positive and negative sample ratios of the two datasets. On the T20H dataset, TED achieves the best results for F1 score and accuracy when the positive sample ratio in the training set is 50\%. The worst results were obtained for F1 score and accuracy when the positive sample ratio in the training set is 20\%. However, the worst F1 score and accuracy of TED are only 3.53\% and 3.01\% lower than the best TED results, respectively. In the T15S dataset, TED performance is still very stable except that the positive sample ratio is 10\%. TED performance drops severely when the positive sample ratio is 10\%, but the F1 score and accuracy are still better than the sub-optimal 9.85\% and 6.03\%, respectively. This can prove that TED has a very powerful tax evasion detection performance even when the labels are unbalanced. 

\begin{figure*}[htbp]
\centering
\subfigure[Embedding Dimension in T20H]{
\label{fig:T20H_dim}
\includegraphics[width=3.65cm]{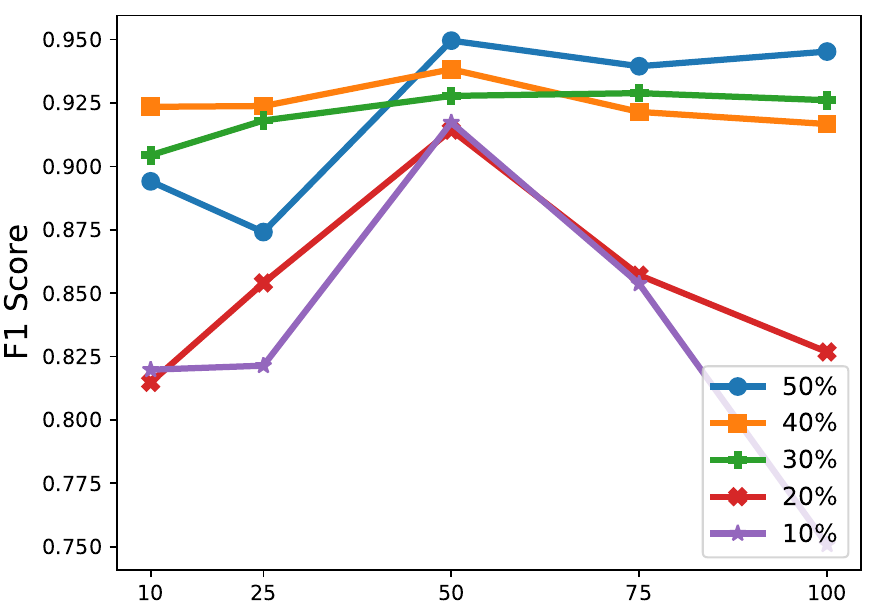}
}
\subfigure[Number of iterations in T20H]{
\label{fig:T20H_ite}
\includegraphics[width=3.65cm]{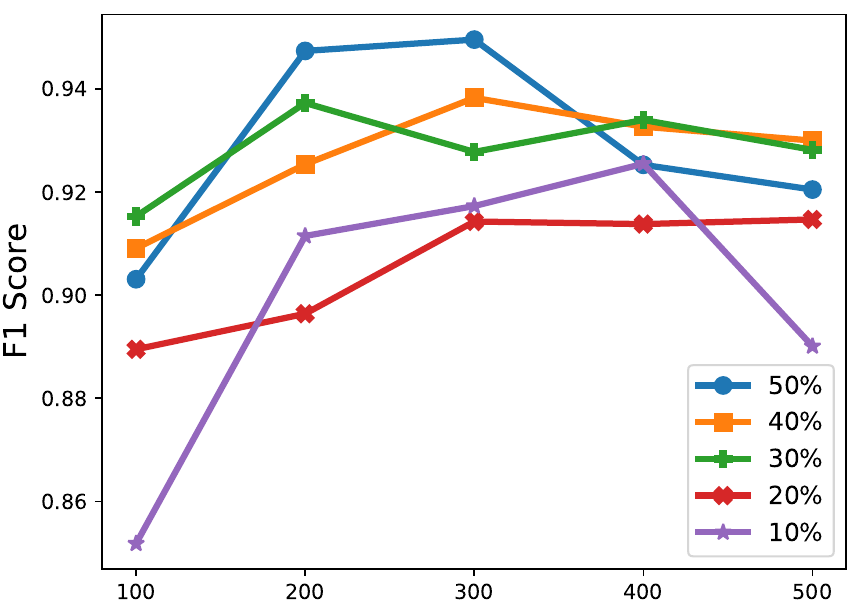}
}
\subfigure[Number of heads in T20H]{
\label{fig:T20H_head}
\includegraphics[width=3.65cm]{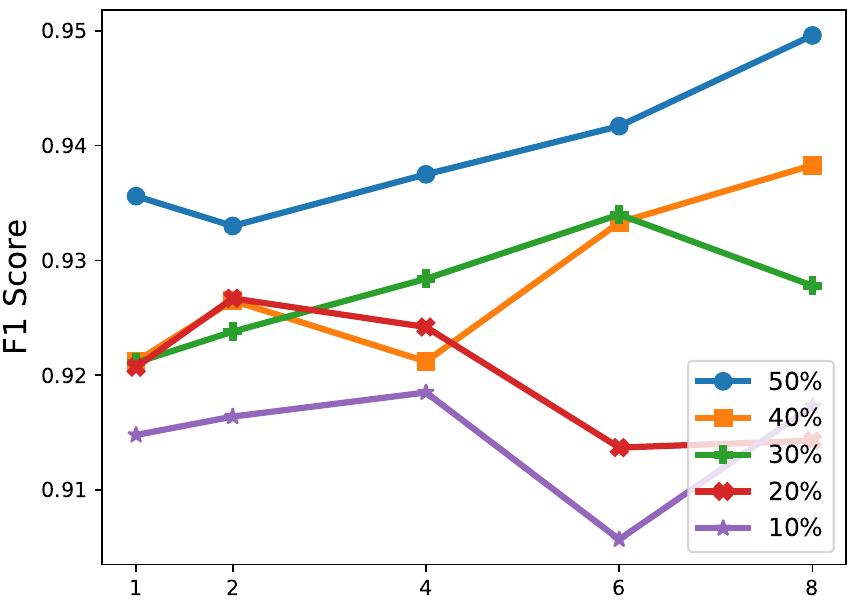}
}
\centering
\subfigure[Embedding Dimension in T15S]{
\label{fig:T15S_dim}
\includegraphics[width=3.65cm]{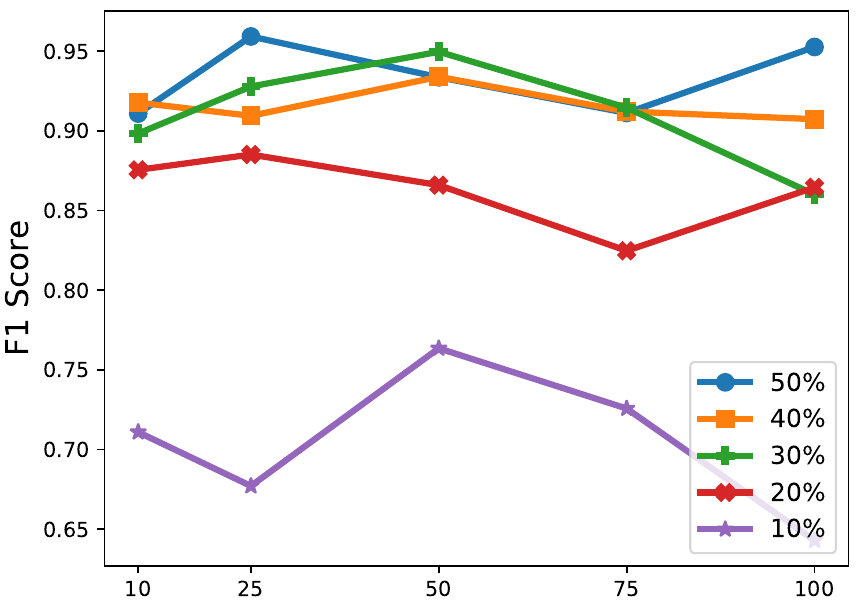}
}
\subfigure[Number of iterations in T15S]{
\label{fig:T15S_ite}
\includegraphics[width=3.65cm]{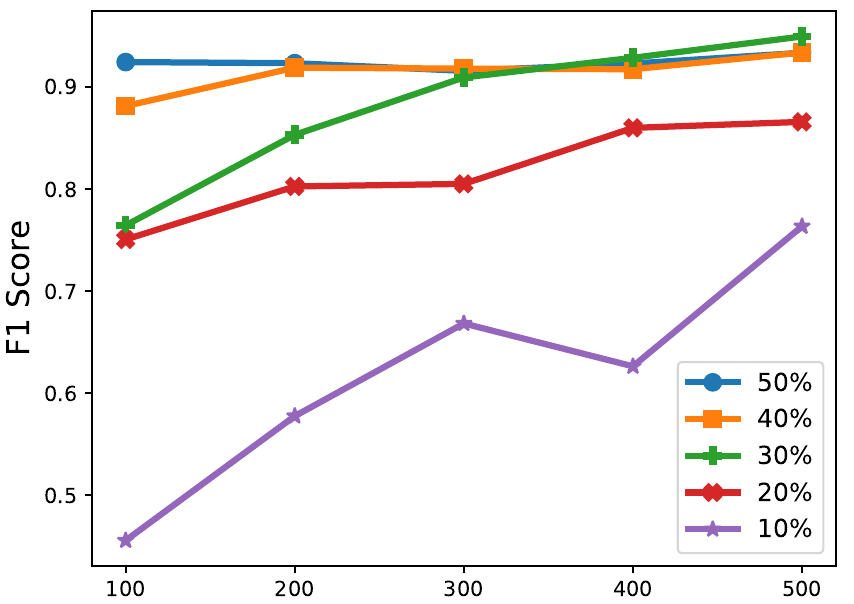}
}
\subfigure[Number of heads in T15S]{
\label{fig:T15S_head}
\includegraphics[width=3.65cm]{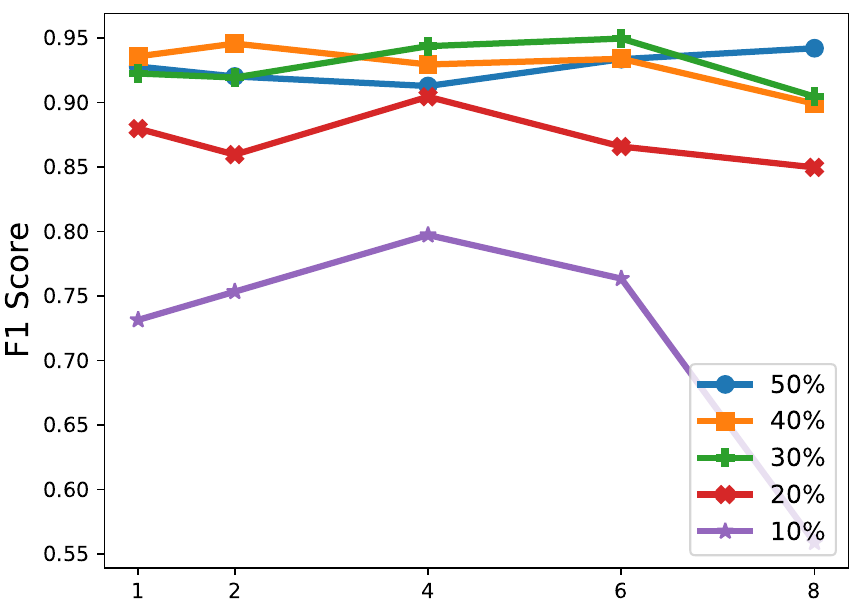}
}
\caption{Sensitivity study of the TED. Sensitivity analysis of TED for different embedding dimensions, number of iterations, and number of heads of TED in two real-world datasets.}
\label{fig:parameters}
\end{figure*}

\subsection{Hyperparameters Sensitivity ($\mathbf{RQ_{3}}$)}
We study the impact of the hyperparameters on TED. We use the F1 score as the evaluation metric and conduct experiments to analyze the results of three parameters (i.e., node embedding dimension, number of iterations, and number of heads) on the two datasets.

$\bullet $
\textbf{Node embedding dimension.} 
The experimental results are shown in Fig.~\ref{fig:T20H_dim} and Fig.~\ref{fig:T15S_dim}. As the dimensionality increases, the classification performance improves first, and then the performance decreases in T20H and T15S. 
This shows that too small of the dimension may only focus on the part of the information, and too large of the dimension may result in overfitting.

$\bullet $
\textbf{Number of iterations.} The experimental results are shown in Fig.~\ref{fig:T20H_ite} and Fig.~\ref{fig:T15S_ite}. As the number of iterations increases, the classification performance first increases and then decreases in T20H. Because the initial model is underfitting, the later model is overfitting. When the epoch is equal to 500, the model has not overfitted on the T15S dataset.

$\bullet $
\textbf{Number of heads.} The experimental results are shown in Fig.~\ref{fig:T20H_head} and Fig.~\ref{fig:T15S_head}. 
As the number of heads increases, the classification performance usually improves,  which can prove that multiple heads are effective. However, on the T15S dataset, F1 decreases when the number of heads is 8, indicating that too many heads may bring noise, and more heads are not always better.


\begin{figure*}[htbp]
\centering
\subfigure[Ablation experiment in T20H]{
\label{fig:T20H_ablation}
\includegraphics[width=5.2cm]{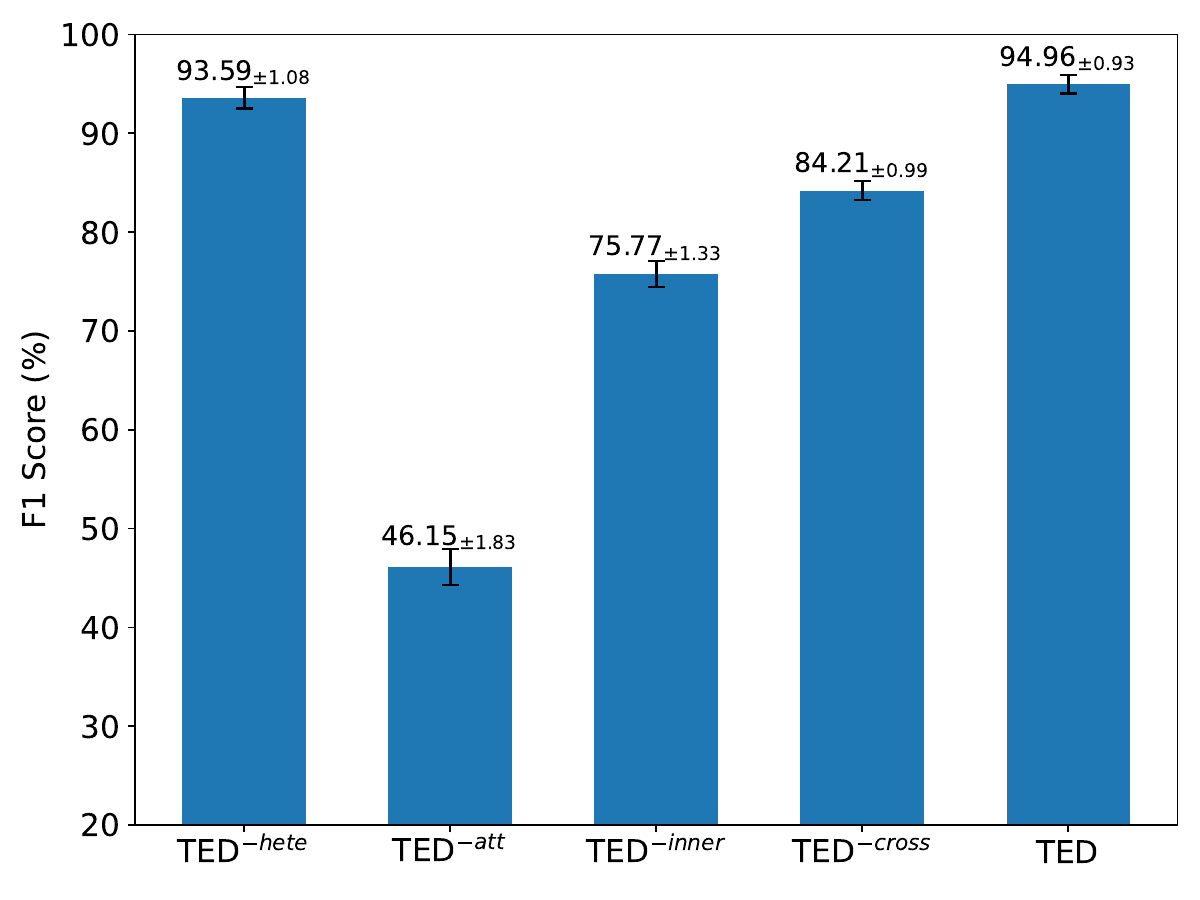}
}
\subfigure[Ablation experiment in T15S]{
\label{fig:T15S_ablation}
\includegraphics[width=5.2cm]{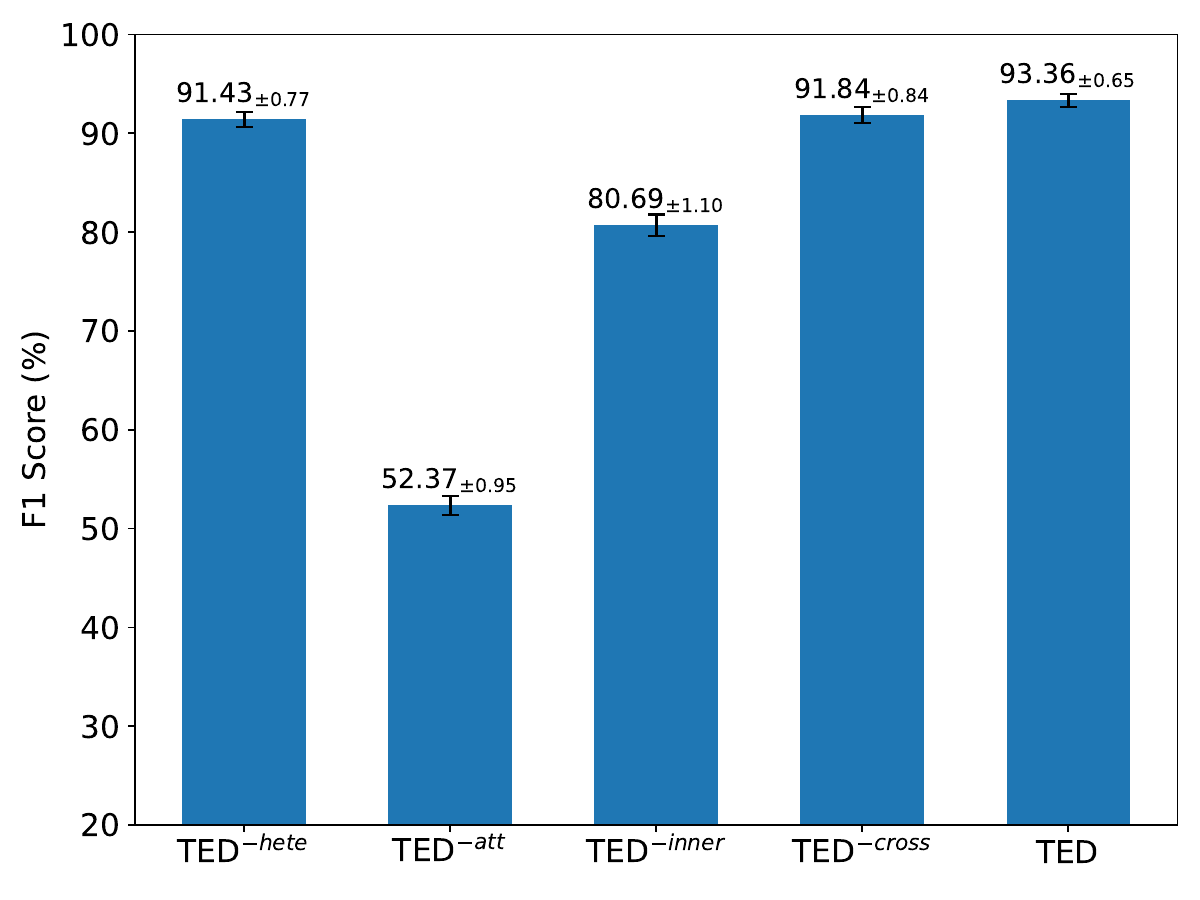}
}
\caption{Ablation experiment of TED in two real-world datasets.}
\label{fig:ablation}
\end{figure*}

\subsection{Ablation Study ($\mathbf{RQ_{4}}$)}
We experiment with different variants of the model to verify the effectiveness of each component of TED. $\mathrm{TED}^{-hete}$ only considers nodes of the company type and ignores nodes of other types at the inner-RPT level. $\mathrm{TED}^{-att}$ is to remove the attention mechanism at TED. $\mathrm{TED}^{-inner}$ and $\mathrm{TED}^{-cross}$ remove the attention mechanism from the inner-RPT level and cross-RPT level, respectively. We use the F1 score as the evaluation metric and conduct ablation experiments on two training sets with a positive sample ratio of 50\%.

Ablation experiment results are shown in Fig.~\ref{fig:T20H_ablation} and Fig.~\ref{fig:T15S_ablation}. Comparing $\mathrm{TED}^{-hete}$ with TED, we can know that only considering company node information and ignoring other types of node information in RPT will affect the detection performance. $\mathrm{TED}^{-inner}$, $\mathrm{TED}^{-cross}$ and TED have significantly improved performance compared to $\mathrm{TED}^{-att}$, and the performance of $\mathrm{TED}^{-inner}$ and $\mathrm{TED}^{-cross}$ is lower than that of TED, which proves the effectiveness of the hierarchical attention model.

\begin{figure*}[htbp]
\centering
\subfigure[DistMult]{
\label{fig:tsne_DistMult}
\includegraphics[width=2cm]{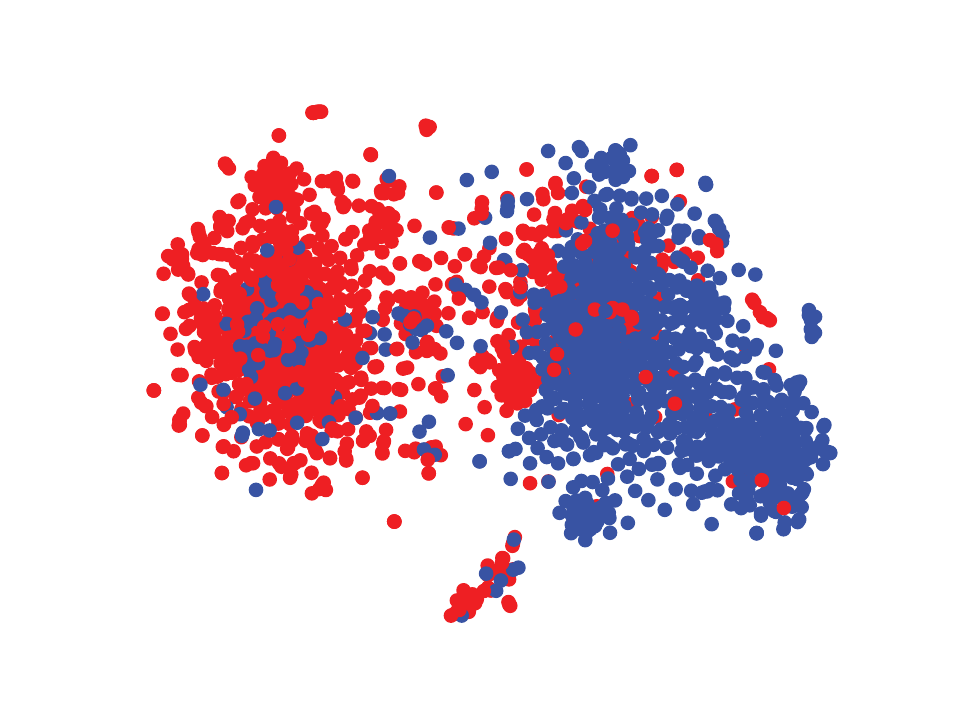}
}
\subfigure[GCN]{
\label{fig:tsne_GCN}
\includegraphics[width=2cm]{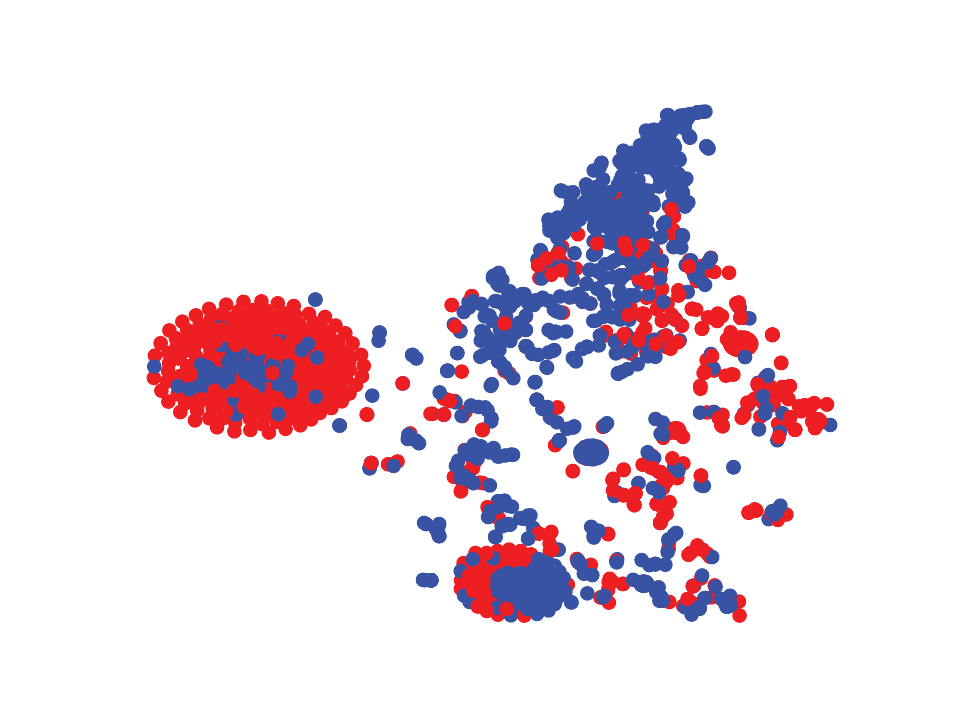}
}
\subfigure[R-GCN]{
\label{fig:tsne_R_GCN}
\includegraphics[width=2cm]{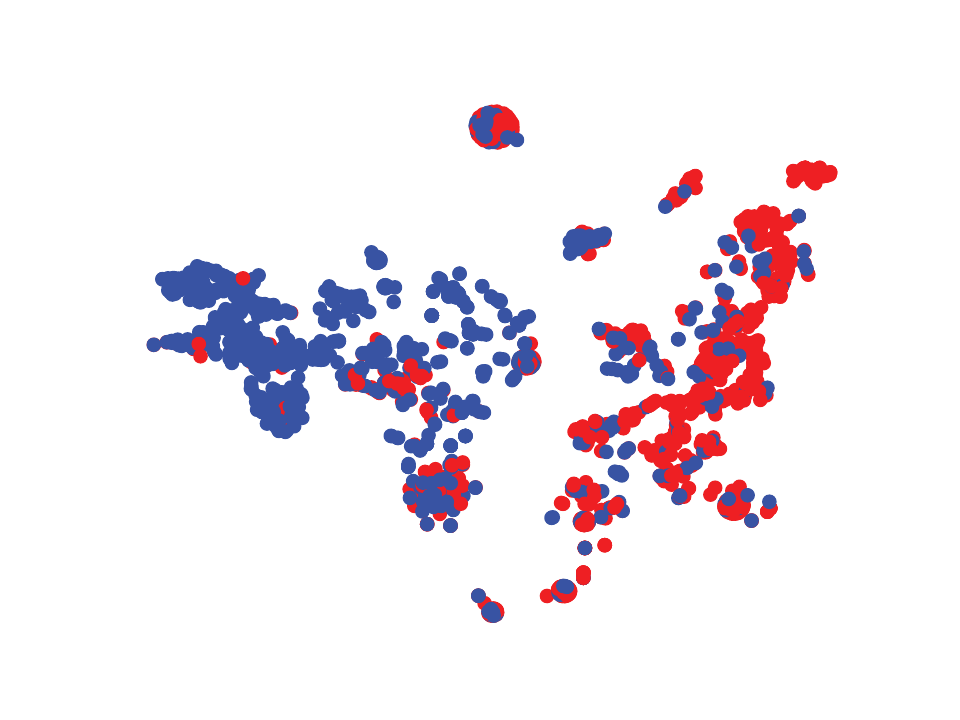}
}
\subfigure[HAN]{
\label{fig:tsne_HAN}
\includegraphics[width=2cm]{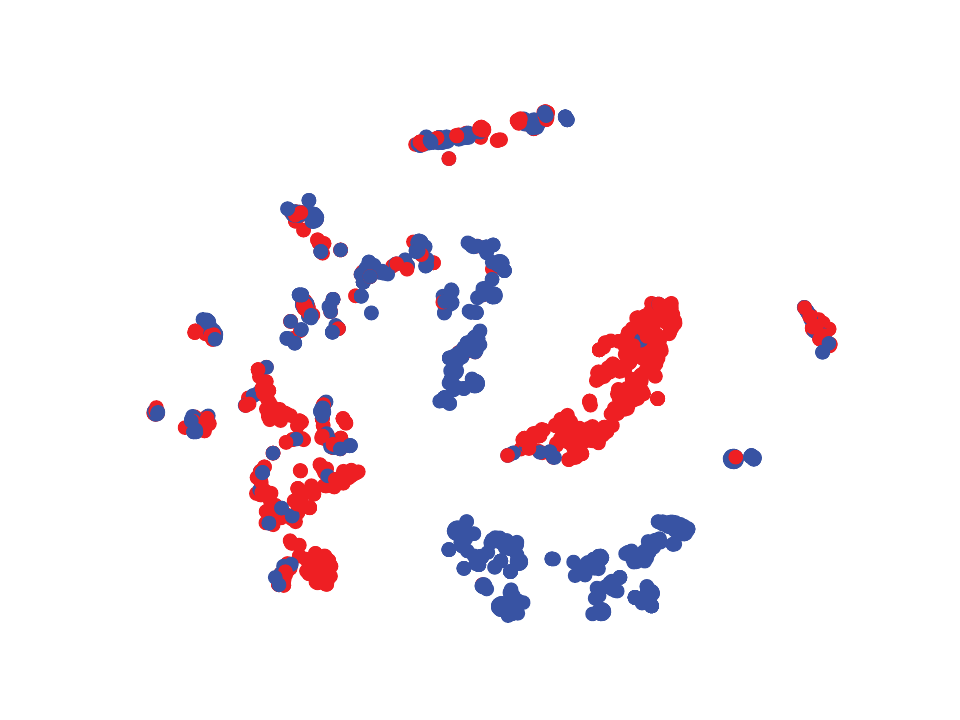}
}
\subfigure[TED]{
\label{fig:tsne_TED}
\includegraphics[width=2cm]{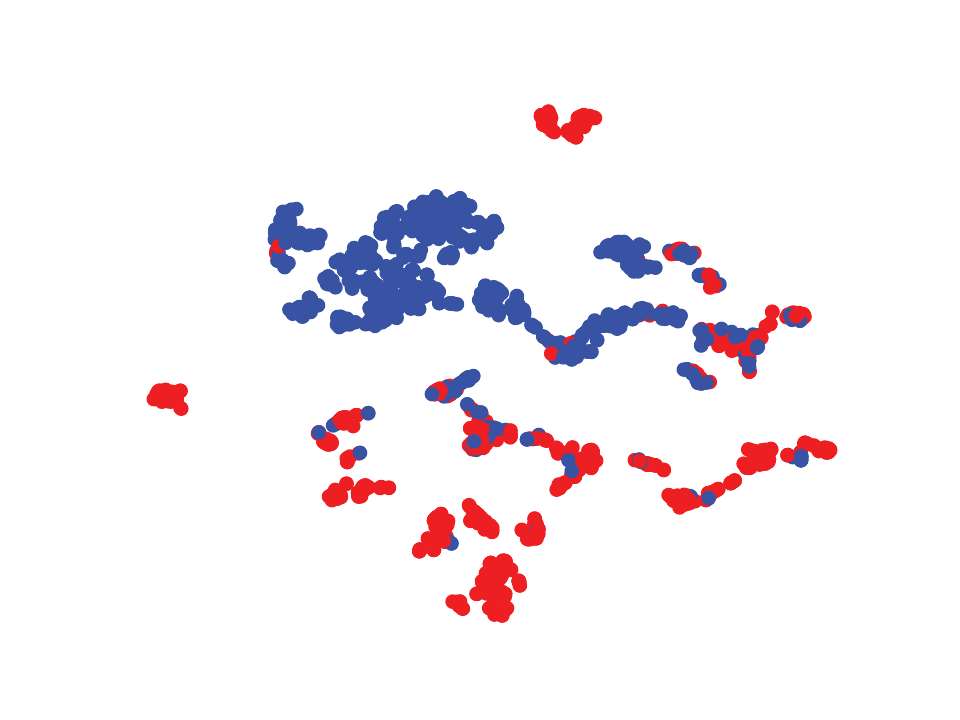}
}
\caption{Visualization embedding on T20H dataset. Each node represents a company. The color of the node represents the type of company, the red nodes represent tax evasion companies, and the blue nodes represent normal companies.}
\label{fig:vis_TSNE}
\end{figure*}

\subsection{Visualization Analysis ($\mathbf{RQ_{5}}$)}
To illustrate the representational capability of our model more intuitively, we use t-SNE~\cite{van2008visualizing} to project the company embeddings learned from the five models (i.e., DistMult, GCN, R-GCN, HAN and our proposed TED) in the T20H dataset into a two-dimensional space, and we use different colors to represent the corresponding company categories, the red nodes represent tax evasion companies, and the blue nodes represent normal companies.

The visualization results are shown in Fig.~\ref{fig:vis_TSNE}, we can find that there is no obvious boundary between the different node categories in the visualization results of DistMult and GCN, and Fig.~\ref{fig:tsne_GCN} shows that GCN may have an over-smoothing problem. R-GCN performs better than GCN because it uses heterogeneous information. Both HAN and TED have better visualization effects. In TED, the embedding distance between normal companies is small. Tax evasion companies retain different tax evasion features and have higher similarities within each cluster. This also reflects that TED can learn better node embedding. 

\begin{figure}[htbp]
\centering
\subfigure[T20H]{
\includegraphics[width=5.2cm]{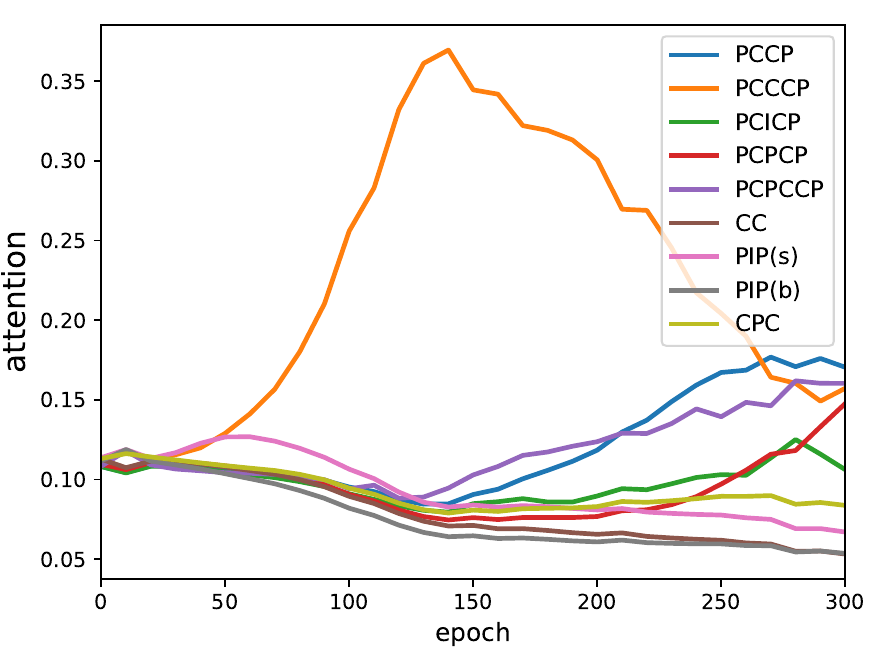}
}
\subfigure[T15S]{
\includegraphics[width=5.2cm]{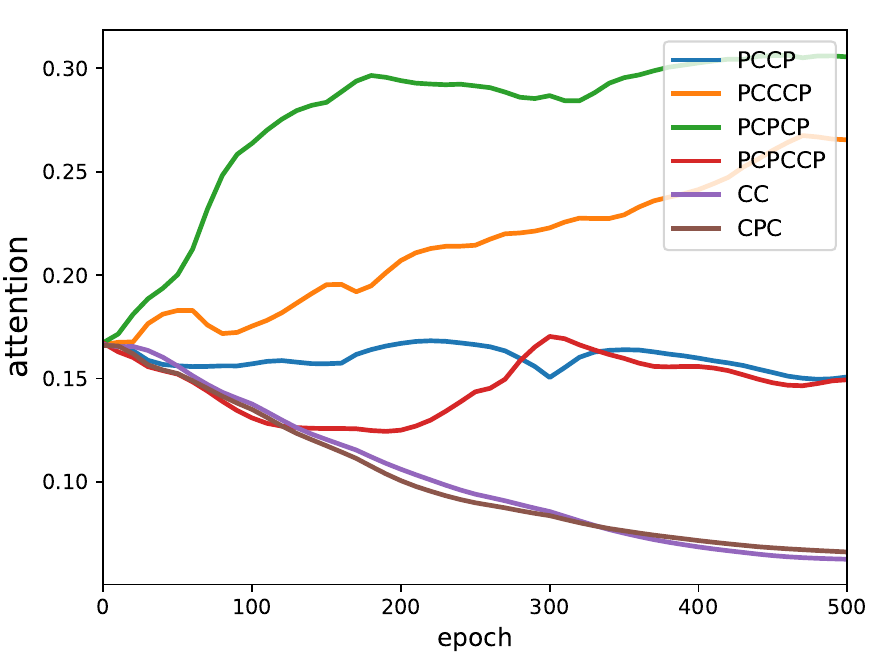}
}
\caption{The changing trends of cross-RPT level attention weight with epochs.}
\label{fig:vis_epoch}
\end{figure}

\begin{figure*}[htbp]
\centering
\subfigure[T20H]{
\label{fig:T20H_ablation}
\includegraphics[width=5.2cm]{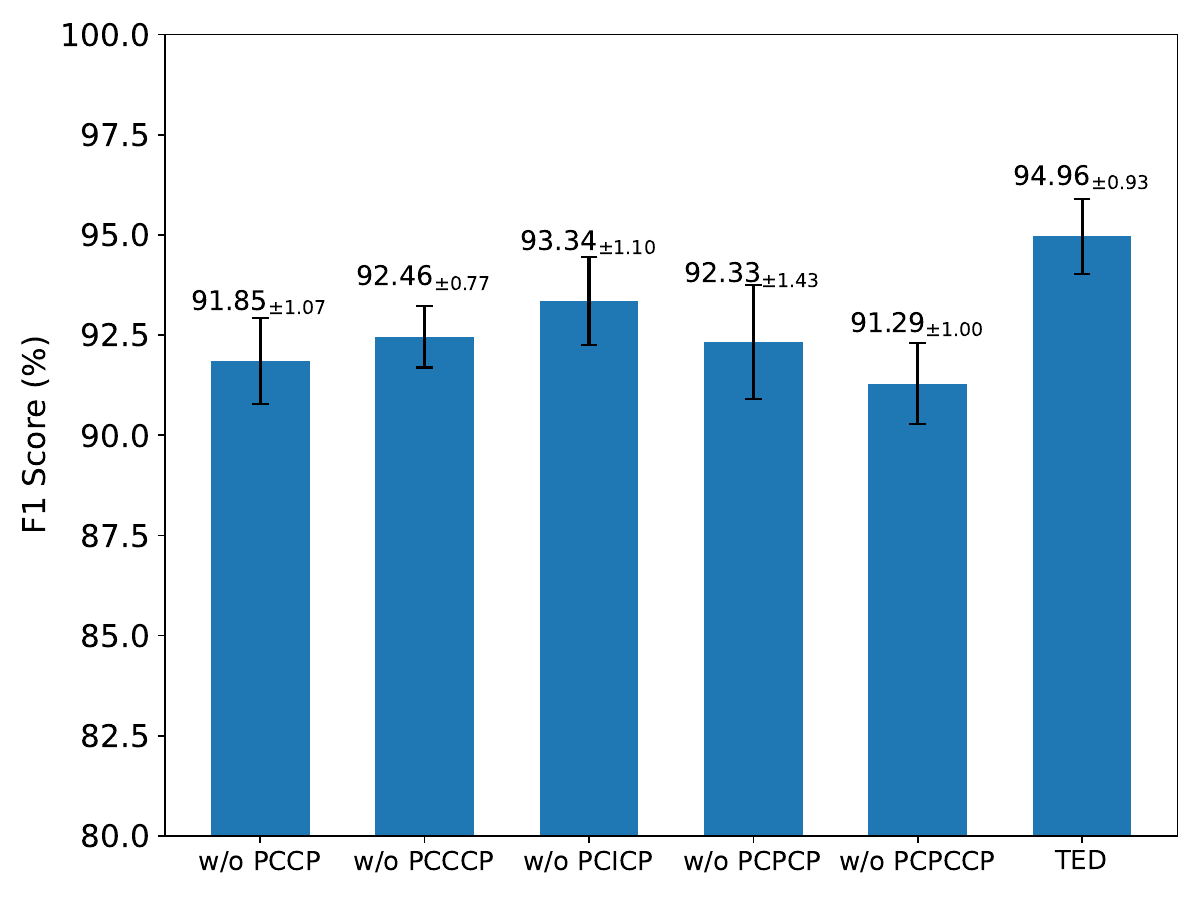}
}
\subfigure[T15S]{
\label{fig:T15S_ablation}
\includegraphics[width=5.2cm]{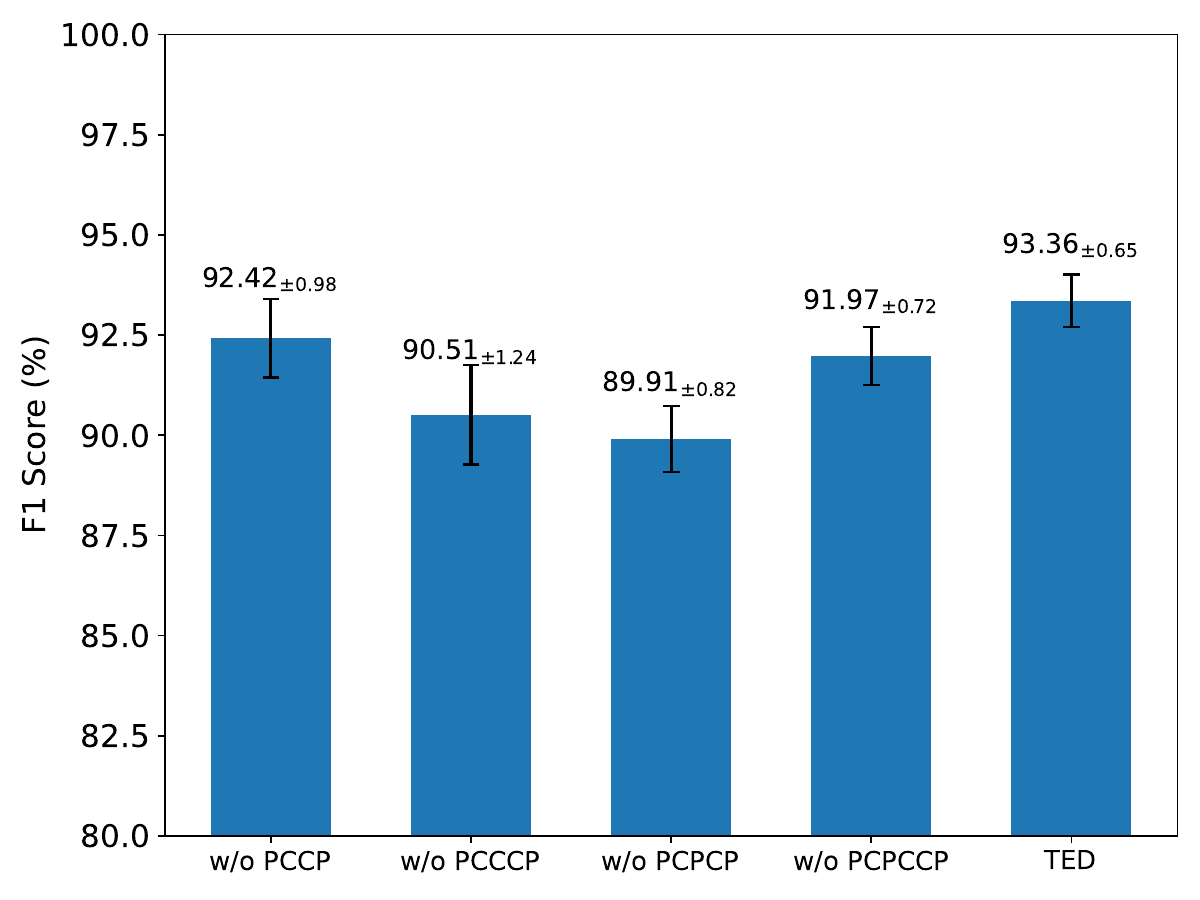}
}
\caption{Evaluate the impact of RPT group selection in two real-world datasets.}
\label{fig:RSA}
\end{figure*}

\subsection{RPT Significance Analysis ($\mathbf{RQ_{6}}$)}
To illustrate the importance of RPT for tax evasion detection, we feed the aforementioned RPT groups ($PCCP$, $PCCCP$, $PCICP$, $PCPCP$ and $PCPCCP$) and also several metapaths ($CC$, $PIP(sell)$, $PIP(buy)$ and $CPC$ for T20H while $CC$ and $CPC$ for T15S) to TED, and visualize the trend of their average attention weights in the cross-RPT level with epoch iterations.

As shown in Fig.~\ref{fig:vis_epoch}, the two datasets originate from different cities and have different distributions of RPT tax evasion behaviors, resulting in different scoring trends. In the same dataset, different RPT scoring trends behave differently due to each RPT contain different topologies and compound relationships. The similarity is that with the increase of epoch, RPT groups gain more attention weights and play a more important role in the final tax evasion detection task. 

Furthermore, we evaluate the impact of RPT group selection on two real datasets. For each evaluation, we remove an RPT group, such as w/o $PCCP$ indicating that there is no $PCCP$ in the input of TED, and assess the significance of that particular RPT group. The experimental results are shown in Fig.~\ref{fig:RSA}. First, we observe that removing any RPT group causes performance degradation. This highlights the significance of each RPT group in achieving optimal performance. Secondly, we discover a positive correlation between the degree of performance degradation and the final attention weights depicted in Fig.~\ref{fig:vis_epoch}. For example, the contribution of $PCCP$ and $PCPCCP$ to the performance is relatively large in T20H, whereas the lack of $PCCCP$ and $PCPCP$ performance degrades severely in T15S. This further shows that even though there are differences in tax evasion patterns and distribution in different regions, TED can better capture information potentially valuable for detection through the attention mechanism.

\begin{figure}[htbp]
\centering
\subfigure[F1 score with Training time]{
\label{fig:training_time}
\includegraphics[width=5.2cm]{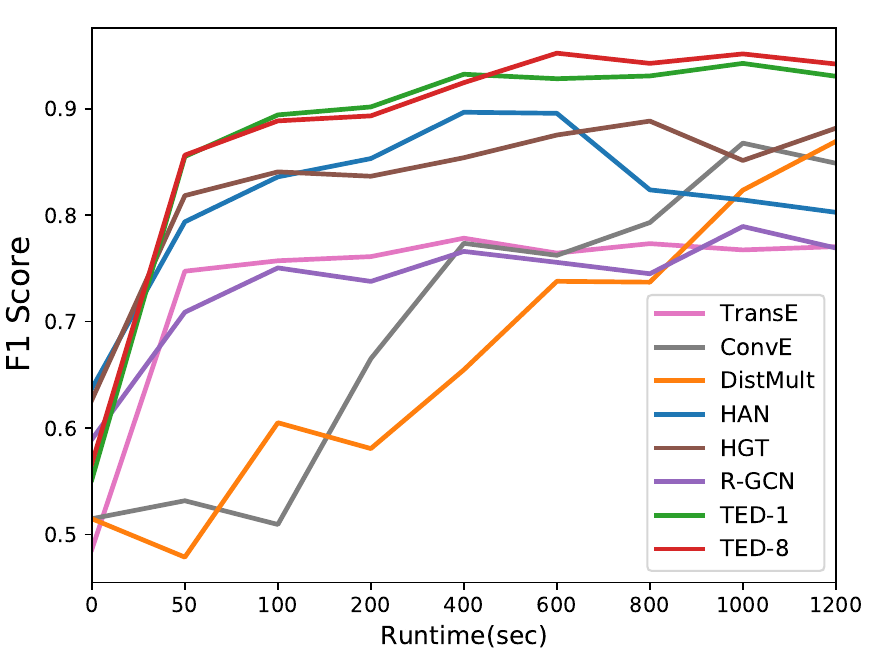}
}
\subfigure[TED convergence time with graph scale]{
\label{fig:convergence_time}
\includegraphics[width=5.2cm]{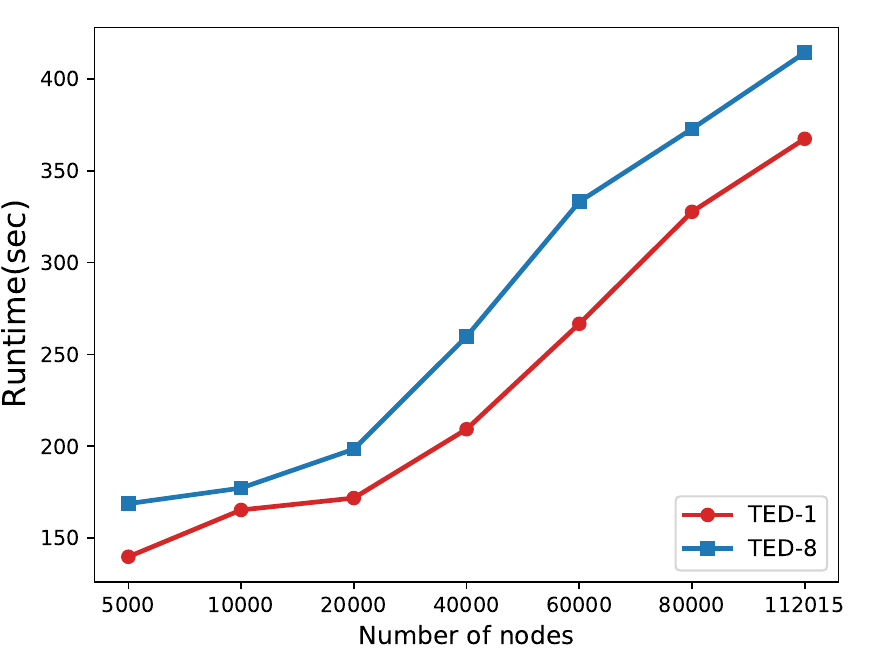}
}
\caption{The efficiency and scalability of TED studied on the T20H dataset. (a) F1 scores of different models with the training time increases; (b) convergence time of TED with the graph scale increases.}
\label{fig:vis_eff}
\end{figure}

\subsection{Efficiency and Scalability ($\mathbf{RQ_{7}}$)}
In this subsection, we study the efficiency and scalability of TED. We repeat all experiments five times and report the mean values in Fig.~\ref{fig:vis_eff}. The experimental results of T15S and T20H are similar, so we only show the results of the T20H dataset. 

We first compare the tax evasion detection performance of 8 models with different training times. More specifically, we calculated the F1 scores of TransE, ConvE, DistMult, HAN, HGT, R-GCN, TED-1 (the inner-RPT level has only one feature extraction head) and TED-8 (the inner-RPT level has eight feature extraction heads) at training times of 0, 50, 100, 200, 400, 600, 800, 1000, and 1200 seconds. The results are shown in Fig.~\ref{fig:training_time}. 
As can be seen, TED-8 performs better than TED-1 since it has more proper feature extraction heads. But the cost is that TED-8 takes more time in each epoch, that's why its F1 is lower than TED-1 in the first 400 seconds. Moreover, TransE and R-GCN converge faster, while ConvE and DistMult take a long time to converge. When TED-1 and TED-8 train for 50 seconds, the F1 score is equivalent to ConvE for 1000 seconds, DistMult for 1000 seconds, HAN for 200 seconds, and HGT for 600 seconds. The above results show that TED takes the shortest time to obtain comparable tax evasion detection performance than others, which proves the efficiency of TED. 

In addition, we study the scalability of our model. Based on the T20H dataset, we construct 7 heterogeneous tax information networks with different scales (the number of nodes is 5000, 10000, 20000, 40000, 60000, 80000 and 112015, respectively) and calculate the model convergence time of TED. 
As shown in Fig.~\ref{fig:convergence_time}, we can find a linear correlation between the convergence time of the TED and the scale of the graph. This results prove that it is possible for TED to detect tax evasion on larger scale data.

\subsection{Generality Analysis ($\mathbf{RQ_{8}}$)}
We analyzed the existing work~\cite{lin2020taxthemis, didimo2020combining, gonzalez2021identifying} in different regions (e.g., China, Italy and Spain), and their datasets could be constructed as heterogeneous graphs. Additionally, some areas may miss some data due to privacy or other issues. However, it can be seen from Table~\ref{tab:SD} and Table~\ref{tab:ncTax20H} that our model still outperforms baselines in the T15S dataset which is missing two node types and four edge types. It can prove that even if part of the data is inadequate, TED is relatively effective, powerful and general. 

\section{Conclusion}
\label{sec:conclusion}
Tax evasion is a serious economic crime that urgently needs to be solved. As far as we know, we first study the tax evasion detection problem under the heterogeneous graph model. In this paper, we introduce a novel algorithm called TED. 
TED uses the RPT group as high-order proximity to filter low-order noisy information, and proposes a novel hierarchical attention model to learn the compound influences from interactive relationships in tax scenarios.
We apply our method to the real risk management system of the tax bureau in China, and conduct extensive experiments on two human-labeled real-world tax datasets, and the results show that our model outperforms the state-of-the-art models on tax evasion detection task.

In future work, we will try to use reinforcement learning to autonomously select meaningful RPT groups and use the performance of downstream tasks to optimize the model. Therefore, the participation of domain experts is entirely unnecessary, which is convenient for application in various fields or more complex heterogeneous graphs.

\bmhead{Acknowledgments}
This research was partially supported by the National Science Foundation of China, No. 62050194, 62002282, 62192781, and 61721002, the China Postdoctoral Science Foundation No. 2020M683492, the MOE Innovation Research Team No. IRT\_17R86, Project of XJTU Undergraduate Teaching Reform No. 20JX04Y and Project of XJTU-SERVYOU Joint Tax-AI Lab.

\bibliography{full_format_references} 




\end{document}